\documentclass{article} 
\usepackage{iclr2025_conference,times}

\usepackage{amsmath,amsfonts,bm}

\def\eqref#1{equation~\ref{#1}}

\def\1{\bm{1}}

\DeclareMathAlphabet{\mathsfit}{\encodingdefault}{\sfdefault}{m}{sl}
\SetMathAlphabet{\mathsfit}{bold}{\encodingdefault}{\sfdefault}{bx}{n}

\usepackage{hyperref}
\usepackage{url}
\usepackage{microtype}
\usepackage{graphicx}
\usepackage{subfigure}
\usepackage{booktabs} 
\usepackage{enumitem}
\usepackage{multirow}
\usepackage{hyperref}
\usepackage{amsmath}
\usepackage{amssymb}
\usepackage{mathtools}
\usepackage{amsthm}
\usepackage[capitalize,noabbrev]{cleveref}

\title{Low Compute Unlearning via Sparse Representations}

\author{Vedant Shah \thanks{Corresponding Author: \texttt{vedantshah2012@gmail.com}} \\
Mila, Universit\'{e} de Montr\'{e}al\\
\And
Frederik Tr\"{a}uble \\
MPI, T\"{u}bingen \\
\And
Ashish Malik \\
Oregon State University \\
\And
Hugo Larochelle \\
Mila, Google DeepMind \\
\And
Michael Mozer \\
Google DeepMind \\
\And
Sanjeev Arora \\
Princeton University \\
\And
Yoshua Bengio \\
Mila, Universit\'{e} de Montr\'{e}al\\
\And
Anirudh Goyal \\
Google DeepMind \\
}

\iclrfinalcopy 
\begin{document}

\maketitle

\begin{abstract}
Machine \emph{unlearning}, which involves erasing knowledge about a \emph{forget set} from a trained model, can prove to be 
costly and infeasible using existing techniques. We propose a 
low compute unlearning technique based on a
discrete representational bottleneck. We show that the proposed
technique efficiently unlearns the forget set and incurs negligible damage to the model's performance on the rest of the data set. We evaluate the proposed technique on the problem of \textit{class unlearning} using four datasets: CIFAR-10, CIFAR-100, LACUNA-100 and ImageNet-1k. We compare the proposed technique to SCRUB, a state-of-the-art approach which uses knowledge distillation for unlearning. Across all three datasets, the  proposed technique performs as well as, if not better than SCRUB while incurring almost no computational cost.\looseness=-1
\end{abstract}

\section{Introduction}
Machine Unlearning \citep{7163042, nguyen2022survey, Zhang2023, xu2023machine, kurmanji2023towards, warnecke2021machine} may be defined as the problem of removing the influence of a subset of the data on which a model has been trained. Unlearning can be an essential component in addressing several problems encountered in deploying deep-learning approaches in practical scenarios. Neural networks such as Large Language Models (LLMs), trained on massive amounts of commonly available data, can exhibit harmful behaviors in the form of generating misinformation, demonstrating harmful biases, or other undesirable characteristics. A major culprit behind these behaviors is the presence of biased or corrupted instances in the training data of these models. To ensure safe model deployment, it is necessary to remove these instances. Another reason to remove instances and make a model behave as if it had not been trained on certain data is concerns about data privacy and the right of end users to expunge their data \citep{MANTELERO2013229, 10.1007/978-3-030-71782-7_35}. For example, an individual might want their data removed from a face recognition system that was trained on their faces such that it is no longer able to identify them. Several regulations are being put in place in order to safeguard the "right to be forgotten" \citep{pardau2018california, inproceedings}. All the above problems can be addressed by unlearning ther specific subset of the training data which gives rise to the harmful behavior of the model in the former cases and an individual's private data in the latter cases.  Apart from these concerns, unlearning can also serve other purposes such as removing outdated data from a model to free the network capacity for more recent or relevant data. With increasing concerns about AI safety and the increasing ubiquity of deep learning models in real-world applications, the problem of unlearning is of critical importance.

\begin{figure*}[ht!]
    \centering
    \includegraphics[width=0.99\linewidth]{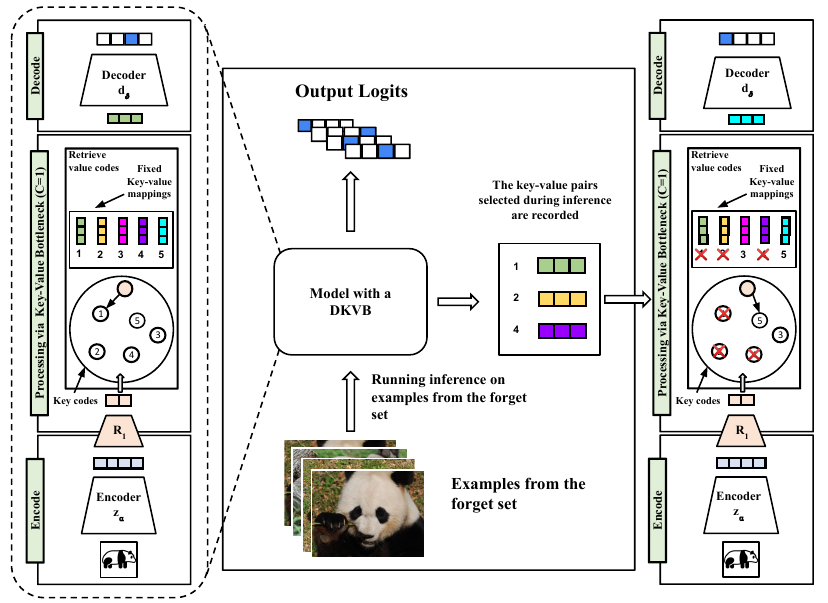}
    \vspace{-0.3cm}
    \caption{A summary of the proposed unlearning approach. \textbf{Left}: The structure of a key-value bottleneck. The encoder is frozen and pre-trained and $R_1$ is a random projection matrix. The values corresponding to the selected keys are retrieved to be used by the decoder. The gradient is backpropagated through the decoder into the values during training. The figure depicts the case with 1 codebook in the DKVB. However, in practice we use multiple codebooks. \textbf{Center:} Examples from the forget set are passed through the trained model and the key-value pairs selected during the forward pass are recorded. \textbf{Right:} The recorded key-value pairs are then masked from the bottleneck. As a result, the key selection is redirected to other keys, with non-informative corresponding values leading to other prediction.}
    \label{fig:1}
    \vspace{-0.4cm}
\end{figure*}

The main challenge in unlearning is maintaining the performance of the model on the data that needs to be retained, called the \textit{retain set}, while unlearning the \textit{forget set}. The naive way to ensure that a model has no information about the forget set is to train from scratch on the retain set. Unlearning techniques aim to achieve the same goal but at a much lower computational cost compared to full retraining. Unlearning in a pretrained network is difficult,  especially in densely connected neural networks, since the value of one parameter may affect the output for all the input examples given to the neural network. A possible solution is to fine tune the model we wish to unlearn only on the retain set. While this would ensure that the performance of the model on the retain set is maintained, it can be computationally infeasible in practice. Other more effective solutions include retraining the model on the training data with a negative gradient for the forget set \citep{golatkar2020eternal, kurmanji2023towards}, or using 
knowledge-distillation-based training objectives to capture information about the retain set while filtering out information about the forget set \citep{kurmanji2023towards, chundawat2023zero}. Nevertheless, all of these approaches require some form of substantial additional compute in order to facilitate unlearning. Moreover, some of the existing approaches additionally require access to the original training data to facilitate unlearning, which may not be possible in many practical applications, e.g., for a model in production which is being trained online on an incoming data stream. The use of large models is becoming more popular and prevalent with the advent of general purpose transformer models. The requirement for additional compute can quickly become impractical in the context of these large models, especially in cases where a model is deployed and needs to be redeployed as quickly as possible after making the necessary changes.

In this article, we argue that specific kinds of \emph{discrete neural information bottlenecks} are highly suited for very efficient and specific unlearning.
Neural information bottlenecks have emerged as useful components in neural network architectures, providing numerous benefits such as improving out-of-distribution (OOD) generalization capabilities and robustness to noisy data \citep{goyal2021coordination, jaegle2021perceiver, liu2021discrete, liu2023stateful}, facilitating large scale unsupervised pre-training and generative modeling \citep{esser2021taming, oord2017neural}, and more recently, helping in continual learning \citep{trauble2023discrete}. In particular, we build  upon \textbf{Discrete Key-Value Bottleneck} (DKVB) proposed in \citet{trauble2023discrete}. DKVB induces sparse representations in the form of key-value pairs which are trained in a \textit{localized and context-dependent manner}. Since these representations are sparse, we hypothesize that it is possible to remove the information about a subset of the training data without damaging the information about the rest of the data---the primary desiderata for a useful unlearning method. Moreover, since the representations are discrete, this may be achieved without requiring any additional compute in the form of retraining or fine tuning, by directly intervening on individual representations.\looseness=-1

We investigate the above-mentioned idea of low compute unlearning in the Discrete Key-Value Bottleneck. Specifically, we focus on the problem of  \textit{class unlearning} in multi-class classification tasks, where the aim is to remove information about a specific class, called the \textit{forget class}, from a trained model. We use the term \textit{retain classes} to refer to the classes other than the forget class that are present in the training data. More specifically, we wish to remove the \textit{influence of the forget class} on the model. We measure this influence using the performance of the model on held-out test datasets corresponding to the forget class and the retain classes.

We propose two approaches for compute efficient unlearning in DKVB - \textit{Unlearning via Examples} and \textit{Unlearning via Activations}. We show that the proposed methods achieve unlearning of the forget class while incurring negligible damage to the model's performance on the retain classes. We compare the proposed methods to SCRUB \citep{kurmanji2023towards}, a recent state-of-the-art approach that requires additional compute to unlearn, on four datasets: CIFAR-10, CIFAR-100, LACUNA-100 and ImageNet-1k. The novelty of our work lies in the largely under-explored idea of using a model architecture with \textit{inherent sparse representations} for unlearning.

\section{Related Work}
The problem of unlearning has been studied in different forms for over two decades. Early works such as \cite{10.1145/2623330.2623661}, 
\citet{cauwenberghs2000incremental} and \citet{Duan2007DecrementalLA} study the problem of \textit{decremental learning} in linear models, where a small number
of samples need to be removed from a model. \citet{ginart2019making} considers unlearning as a problem of deleting individual data points from a model. They give a
probabilistic definition, formalize the notion of efficient data deletion, and propose two deletion efficient learning algorithms. \citet{guo2019certified} 
introduces \textit{certified removal} - a theoretical guarantee of indistinguishability between a model from which data was removed and a model that never saw the
data. \citet{izzo2021approximate} distinguishes between exact unlearning and approximate unlearning and proposes a compute-efficient  approximate data deletion 
method, and a new metric for evaluating data deletion from these models. \citet{golatkar2020eternal} and \citet{kurmanji2023towards} cast unlearning into an 
information theoretic framework. \citet{golatkar2020forgetting} proposes Neural Tangent Kernel (NTK) \citep{jacot2018neural} theory-based approximation of the weights of the unlearned network. Multiple works also delve into the more philosophical, ethical, and legal aspects of unlearning and the 
``right to be forgotten" \citep{Kwak2017LetMU, VILLARONGA2018304}. \citet{chundawat2023zero} and \citet{tarun2023fast} learn error minimization and error maximization-based noise matrices which are used to finetune the trained model in order to do unlearning. \citet{chundawat2023zero} further uses a generator that generates pseudo data points for unlearning in order to operate in a data-free regime.

Most relevant to our work, \citet{kurmanji2023towards} introduces SCRUB, a knowledge distillation-based unlearning method. SCRUB considers the
original model as a teacher model and trains a student model to obey the teacher model on the retain set and disobey it on the forget set. This is done by computing the KL Divergence between the output distributions of the two models and training
the student model to maximize it on the forget set (called a 
\textit{max-step}) and minimize it on the retain set (called a \textit{min-step}). The student model is simultaneously also optimized for minimizing the task loss on the retain set. The training consists of \textit{mstep} \textit{max-steps}. The \textit{max-steps} and \textit{min-steps} are executed alternatively. \citet{Chen2023BoundaryUR}, similarly to us, focuses on class unlearning rather than unlearning specific instances in the data. Unlearning is done by destroying the decision boundary of the forget class. The authors propose two boundary shift methods termed as \textit{Boundary Shrink} and \textit{Boundary Expanding}.

\citet{jia2023model} and \citet{mehta2022deep} investigate unlearning in context of model sparsity. \citet{jia2023model} leverages the Lottery Ticket Hypothesis, \cite{frankle2018lottery} leverages using parameter pruning on a trained dense model to identify the token subnetwork. They observe that applying standard unlearning approaches to a sparsified networks is better as compared to doing unlearning directly on the dense network. \citet{mehta2022deep} identifies the Markovian Blanket of parameters corresponding to the examples to be unlearnt and updates those parameters. Their approach can be seen as applying sparse unlearning updates to the network.

While most of the approaches discussed above improve upon the naive and intractable baseline of retraining on the retain set, all of them require a substantial amount of additional computation in the form of optimizing an objective function for unlearning. This additional compute requirement can quickly become infeasible whenever large models are involved. The proposed approach in this work, on the other hand, requires negligible computation for unlearning. Any computation that may be required is in the form of running inference on the forget set. We point out that sparsity is a critical dimension that determines the effectiveness of unlearning:  extremely sparse representations make unlearning trivial, whereas fully distributed representations intertwine knowledge in a way that makes compute-efficient unlearning a serious challenge. Previous methods studying sparsity in the context of unlearning such as \citet{jia2023model} and \citet{mehta2022deep} propose the use of pruning techniques to first sparsify the network. These approaches start with dense trained models and leverage sparsity for unlearning. In contrast, we propose using sparsity as an in-built inductive bias in the model during the initial training which makes the model suitable for unlearning involving minimal compute requirements. On the other hand, \citet{jia2023model} sparsify the model after it has been trained. \citet{mehta2022deep} involves sparse updates to the model parameters as discussed previously. However, these sparse updates are utilized during unlearning as opposed to during training of the original model in the proposed approach. We identify a sweet spot on the continuum between local and distributed learning that allows for both, compute-efficient unlearning and simultaneously obtaining the same generalization performance.


\citet{xu2023machine} have introduced a taxonomy that categorizes existing research on unlearning based on different approaches and aims. 
In this classification, our methods fall within the \textit{Model Pruning} category by means of disabling specific (key, value) pairs within the bottleneck. Although one could argue that our methods lean towards a \textit{weak unlearning} strategy--given the pre-trained backbone might retain some information about the forget set--our approach deviate from the strict definition of \textit{weak unlearning} as outlined by \citet{xu2023machine}. As an example, when considering a non-parametric decoder, our methods affect intermediate rather than final model activations.



\section{Background and Notations}
\textbf{Unlearning}: Let $\mathcal{D}_{train} = \{x_i, y_i\}^N_{i=1}$ be a training dataset and $\mathcal{D}_{test}$ be the corresponding test dataset. In our experiments, we consider the setting of class unlearning, wherein we aim to unlearn a class $c$ from a model trained with a multiclass classification objective on $\mathcal{D}_{train}$. $c$ is called the \textit{forget class} or the \textit{forget set}. Given $c$, we obtain $\mathcal{D}_{train}^{forget} \subset \mathcal{D}_{train}$ such that $\mathcal{D}_{train}^{forget} = \{(x, y) \in \mathcal{D}_{train} | y = c\}$. The complement of $\mathcal{D}_{train}^{forget}$ is $\mathcal{D}_{train}^{retain}$, i.e., subset of $\mathcal{D}_{train}$ that we wish to retain. Thus $\mathcal{D}_{train}^{retain} \cup \mathcal{D}_{train}^{forget} = \mathcal{D}_{train}$. Similarly, from $\mathcal{D}_{test}$, we have $\mathcal{D}_{test}^{forget} = \{(x, y) \in \mathcal{D}_{test} | y = c\}$ and its complement $\mathcal{D}_{test}^{retain}$. We refer to $\mathcal{D}_{train}^{retain}$ and $\mathcal{D}_{test}^{retain}$ as the retain set training and test data; and $\mathcal{D}_{train}^{forget}$ and $\mathcal{D}_{test}^{forget}$ as the forget set training and test data, respectively.

\textbf{Discrete Key-Value Bottleneck}: A discrete key-value bottleneck (DKVB) \citep{trauble2023discrete} consists of a discrete set of coupled key-value codes. The bottleneck contains $C$ codebooks with each codebook containing $M$ key-value pairs. Models with DKVB use a pre-trained and frozen encoder to encode the input into a continuous representation. This input representation is then projected into $C$ lower dimension heads and each head is quantized to the $top-k$ nearest keys in the corresponding codebook. The values corresponding to the selected keys are averaged, and used for the downstream task. The keys in the codebooks are frozen and initialized to cover the input data manifold whereas the values are learnable. The mapping between the keys and values is non-parametric and frozen. Thus, the gradient is not propagated between the values and keys during training of the model. Since the values are retrieved and updated sparsely, and all the components except the value codes and the decoder are frozen, DKVB stores information in the form of input-dependent, sparse and localized representations (i.e., the value codes). These inductive biases allow the framework to exhibit improved generalization under distribution shifts during training, as shown empirically as well as theoretically in \citet{trauble2023discrete}. Figure \ref{fig:1} (Left) shows an overview of a model with a DKVB where $C=1$, $M=5$ and top-$k = 1$.

\section{Unlearning via Sparse Representations}

\label{sec:method}

\textbf{Learning a Discrete Key Value Bottleneck.} A Discrete Key Value Bottleneck (DKVB) model is first trained on the given dataset using the standard negative log-likelihood (cross-entropy loss) training objective for multi-class classification.  We use a non-parametric average pooling decoder and test the proposed approaches on two pretrained backbones: 1.) a CLIP \citep{radford2021learning} pre-trained ViT-B/32 \citep{dosovitskiy2020image} and 2.) a ResNet-50 pretrained on ImageNet in a supervised fashion. Then we proceed to unlearn a specific subset of data from these models. Before training with the classification objective, we do a \textit{key initialization} for the DKVB models on the same dataset.\looseness=-1

\textbf{Key Initialization in DKVB models.} The mapping between keys and values in the discrete key-value bottleneck is non-parametric and frozen. As a result, there is no gradient (back)propagation from the values to the keys. Thus, it becomes essential for the keys to be initialized before learning the values and decoder, such that they broadly cover the feature space of the encoder. This ensures that the representations are distributed sparsely enough. As in \citet{trauble2023discrete}, we use exponential moving average (EMA) updates \citep{oord2017neural, razavi2019generating} to initialize the keys of the DKVB models. The key-initialization is done on the same train dataset $\mathcal{D}_{train}$ which we want to train the model on. The key initializations depend solely on the input encodings of the backbone and hence do not require access to any labeled data. 

\textbf{Inference for Unlearning.} We propose to achieve unlearning in DKVB models by excluding key-value pairs from the bottleneck such that they cannot be selected again. Numerically, this masking is done by setting the quantization distance of the selected keys to `infinity'. Figure~\ref{fig:1} (center and right column) shows an overview of the proposed methods. More specifically, we experiment with two methods, \emph{Unlearning via Activations} and \emph{Unlearning via Examples}, described as follows. \looseness=-1

\textbf{Unlearning via Examples.} In this method, we analyze the effect of unlearning a subset of $N_e$ examples belonging to the forget set. $N_e$ examples are randomly sampled from the forget set training data ($\mathcal{D}_{train}^{retain}$) and are input into the model having a DKVB. All key-value pairs that are selected during forward propagation across the $N_e$ examples are flagged. These key-value pairs are then masked out from the bottleneck. Technically, this approach requires access to the original training data corresponding to the forget class. However, it is also possible to carry out this procedure with a proxy dataset that has been sampled from a distribution close enough to that of the forget set. This approach is motivated by the assumption that such a dataset would likely utilize roughly the same set of selected key-value pairs. 

\textbf{Unlearning via Activations.} In this second method, we analyze the effect on the quality of unlearning by deactivating different numbers of key-value pairs corresponding to the forget set. We refer to the key-value pairs that have been selected as inputs to the decoder as \textit{activations}. The entire forget set is forward-propagated through the DKVB model and all the key-value pairs selected across all examples of the forget class are recorded. Next, we mask the top-$N_a$ most frequently selected key-value pairs from the bottleneck. The requirement of accessing the original training data for this method can be avoided by caching all the activations corresponding to the forget set during the last epoch of training. Further, similar to the previous case, unlearning via activations may also be performed given access to data that has been sampled from a distribution close enough to the distribution of the forget set. 

Both approaches 
are two different ways of achieving a common objective: to exclude a subset of activations corresponding to the forget set. However, using one approach over the other may be more practical or even necessary, depending on the task at hand.
In both the above approaches, we do not do any form of retraining or fine-tuning. The only computation which may be necessary is incurred during the inference stage for recording the key-value pairs which have been utilized for the forget set. Hence both approaches require negligible additional compute. Moreover, the requirement of access to original training data of the forget class can also be circumvented under appropriate assumptions, making the proposed approaches \textit{zero-shot unlearning methods}.\looseness=-1

\section{Experiments and Results}
\label{sec:expt}

The goal of our experiments is two-fold. First, we validate that proposed methods of the \textit{Unlearning via Activations} and \textit{Unlearning via Examples} in models with a DKVB (Section~\ref{subec:results_zeroshot}), and show that the proposed methods are competitive with the baselines (Section~\ref{sec:compbase}) in unlearning the forget class while incurring minimal damage to the performance of the models on the retain class. Second, we compare the compute efficiency of the proposed methods against that of the baselines. More specifically, we report the number of floating point operations (FLOPs) required during the procedure of unlearning. 
Before presenting these results, we describe our experimental setup.

\subsection{Experimental Setup}
\label{subsec:exp_setup}
\textbf{Benchmark datasets}
We validate the proposed methods using experiments across four base datasets: CIFAR-10 with 10 distinct classes, CIFAR-100 \citep{krizhevsky2009learning} with 100 distinct classes,  LACUNA-100 \citep{golatkar2020eternal} with 100 distinct classes and ImageNet-1k \citep{russakovsky2015imagenet} with 1000 distinct classes. LACUNA-100 is derived from VGG-Faces \citep{Cao18} by sampling 100 different celebrities and sampling 500 images per celebrity, out of which 400 are used as training data and the rest are used as test images. 

\textbf{Models} On the aforementioned three datasets we study the following types of model architectures:
\begin{enumerate}[label=(\alph*), topsep=0pt, itemsep=0ex, partopsep=1ex, parsep=1ex, leftmargin=*]
    \item[(a)] \textbf{Backbone + Discrete Key-Value Bottleneck (Ours)}: Overall, this architecture consists of three components: 1) the frozen pre-trained backbone 2) the Discrete Key-Value Bottleneck (DKVB) and 3) a decoder, as shown in Figure~\ref{fig:1}. For the DKVB, we use 256 codebooks, with 4096 key-value pairs per codebook (approximately 1M pairs overall) as in \citet{trauble2023discrete}. \looseness=-1
    \item[(b)] \textbf{Backbone + Linear Layer (Baseline)}: As a baseline, we replace the Discrete Key Value bottleneck and the decoder in the above model architecture with a linear layer. Thus, the two components of this model are 1) a frozen pre-trained backbone and 2) a linear layer. This model will be used for all the baseline methods. 
\end{enumerate}
In each model, we use a pre-trained frozen CLIP \citep{radford2021learning} ViT-B/32 and ImageNet supervised pre-trained ResNet-50 as our encoder backbones. We refer the reader to the appendix for additional implementation details.

\textbf{Training the Base Models} We then train both model architectures on the full training sets of each dataset.
Since the backbone is frozen, for the baseline models, only the weights of the linear layer are tuned during initial training (and later unlearning). Since we use only one linear layer, we do not do any pre-training (beyond the backbone), unlike in previous works \citep{kurmanji2023towards, golatkar2020eternal, golatkar2020forgetting}.
Table~\ref{tab:init_perf} shows the performance of these trained models on the train and test splits of the complete datasets. 
Starting from these base models trained on the full datasets, we will validate the ability to unlearn previously learned knowledge.
 

\begin{figure*}[t]
    \centering
    \includegraphics[width=0.23\linewidth]{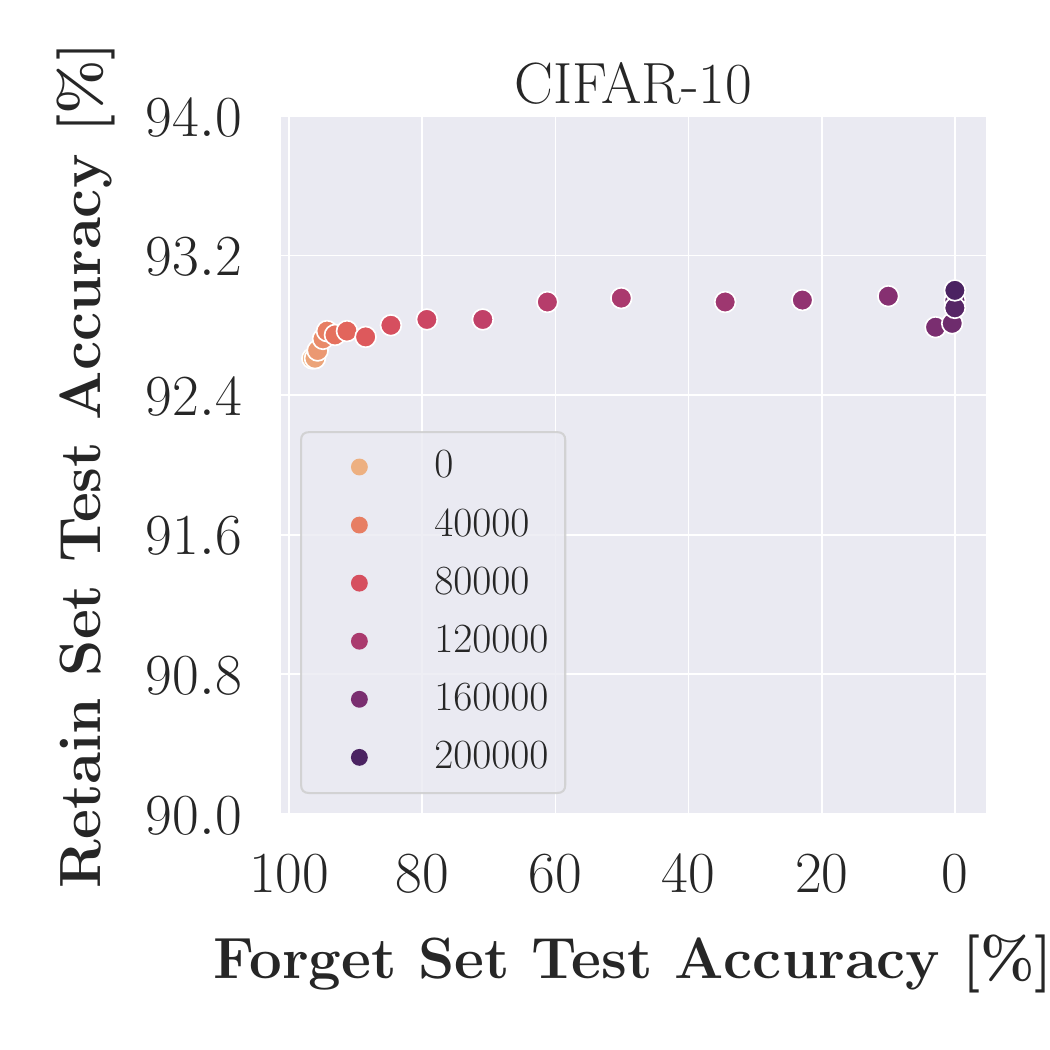}
    \includegraphics[width=0.23\linewidth]{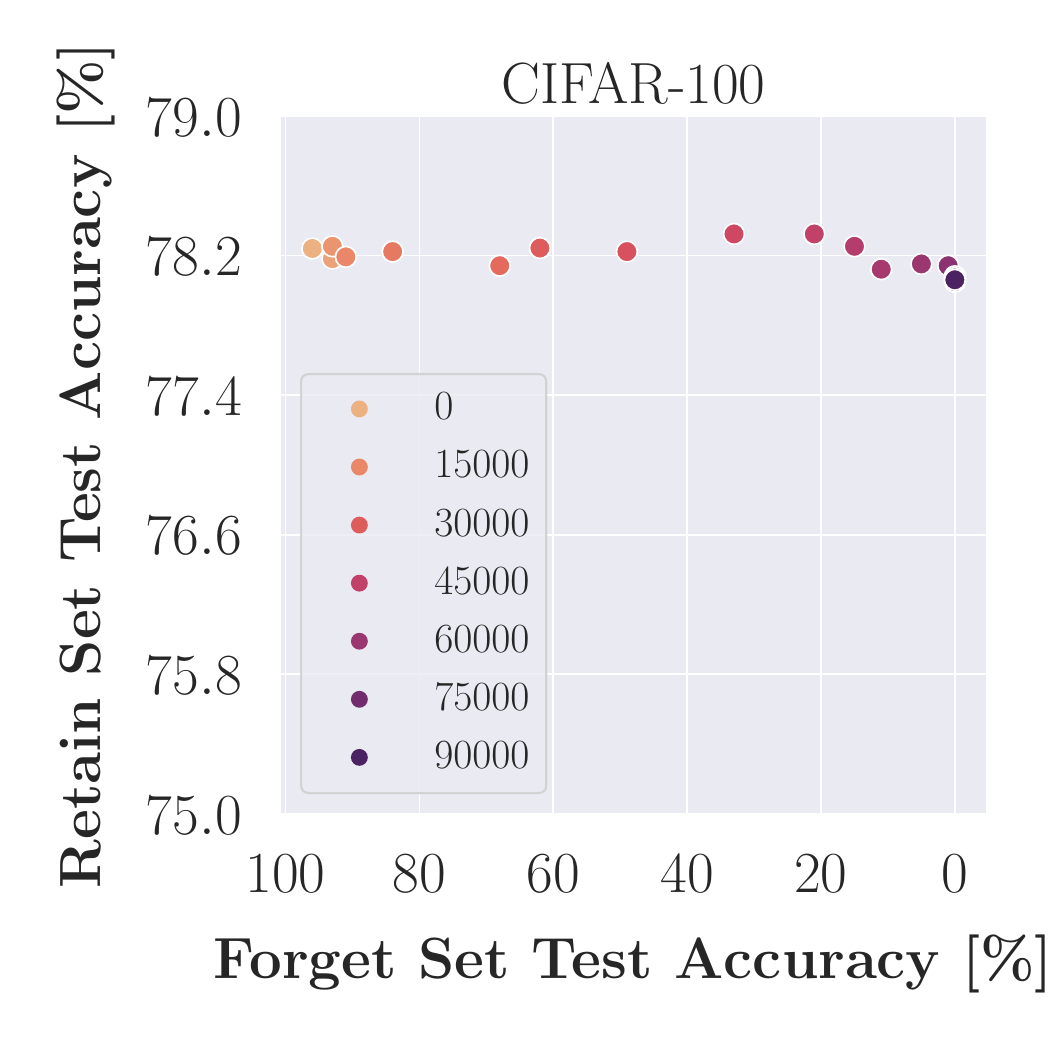}
    \includegraphics[width=0.23\linewidth]{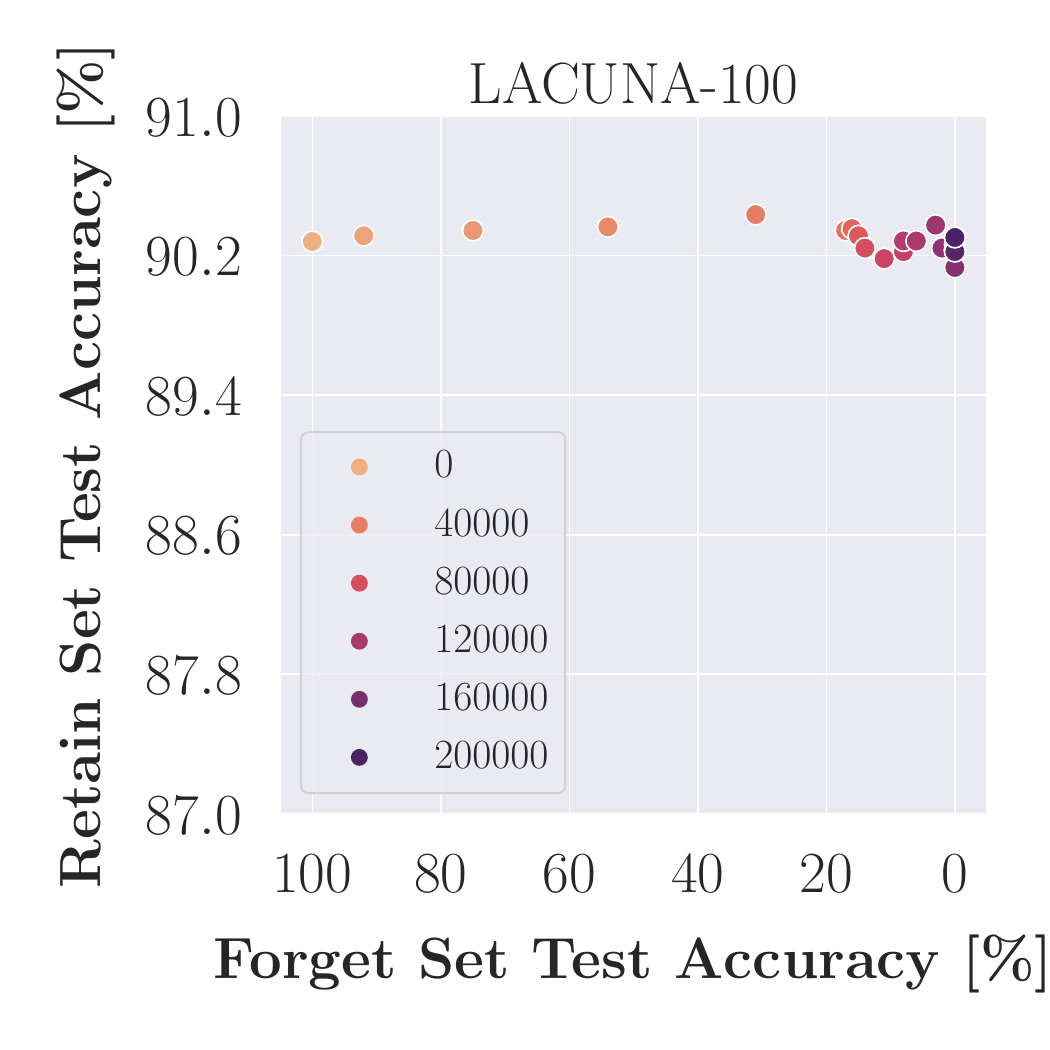}
    \includegraphics[width=0.23\linewidth]{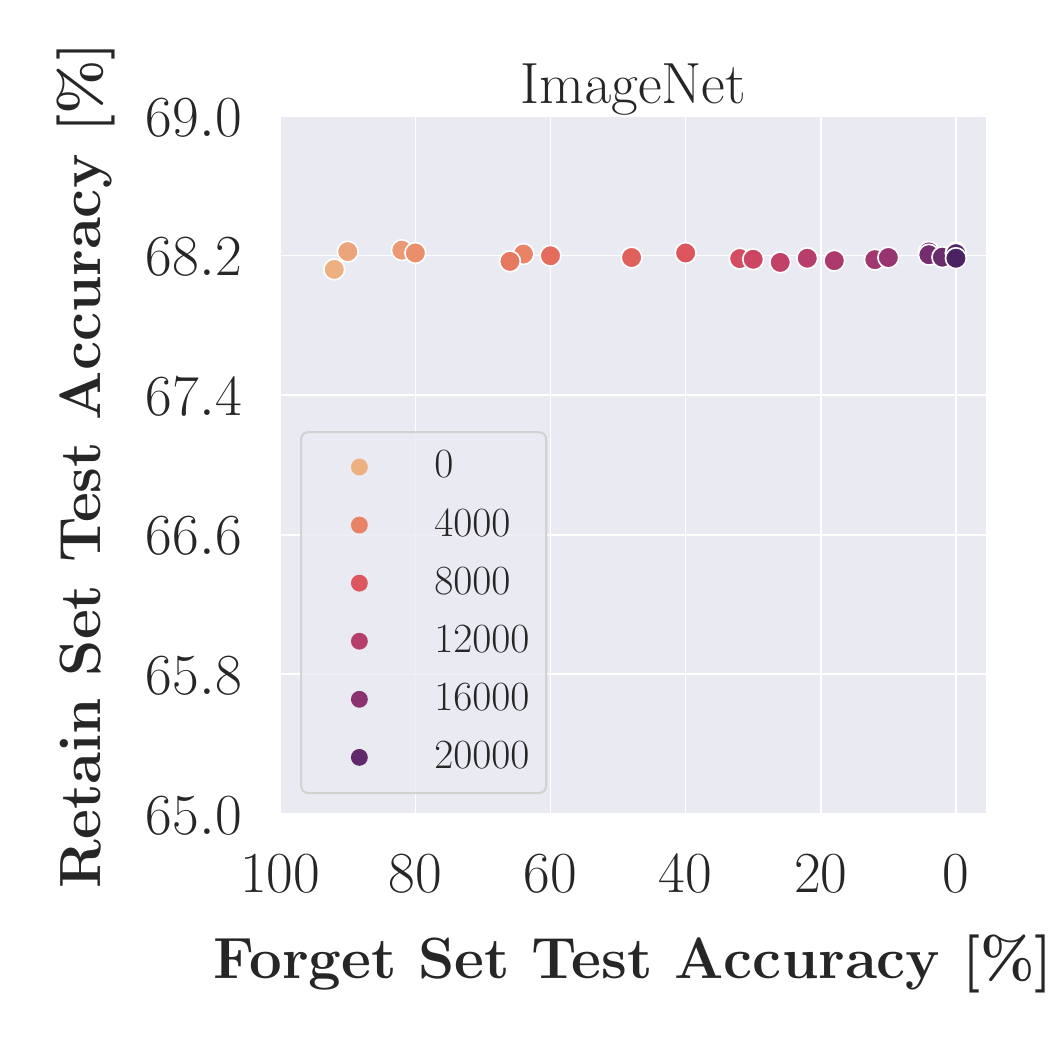}
    \label{fig2:a}
    \\
    \includegraphics[width=0.23\linewidth]{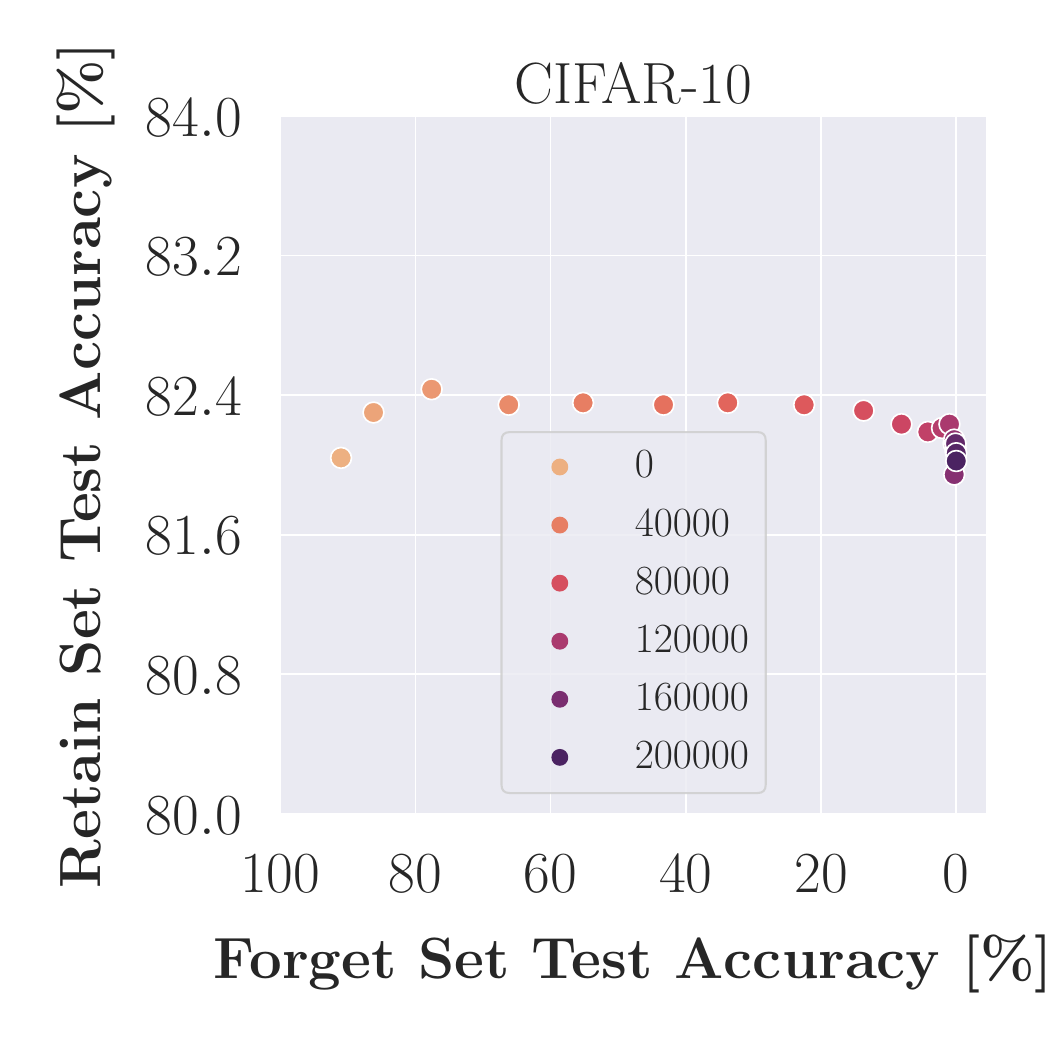}
    \includegraphics[width=0.23\linewidth]{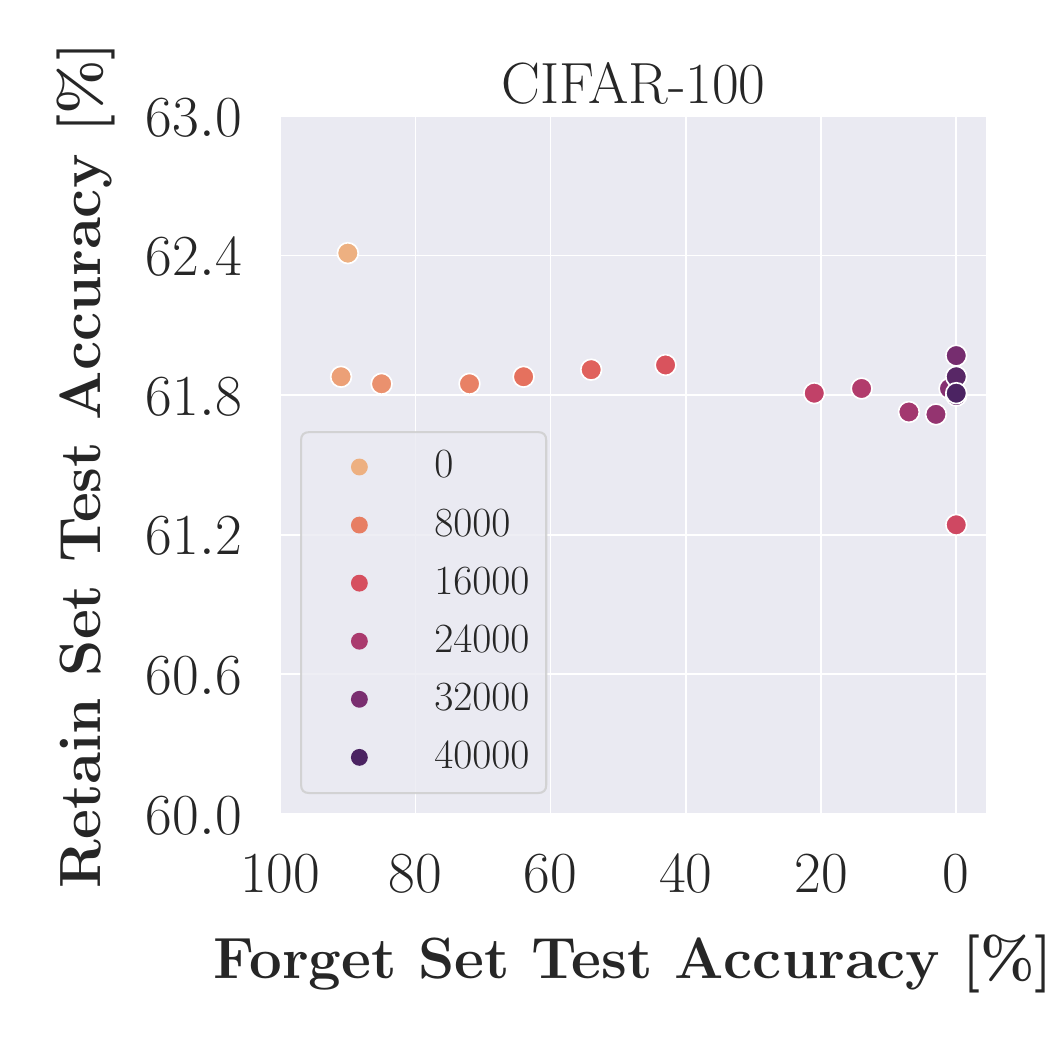}
    \includegraphics[width=0.23\linewidth]{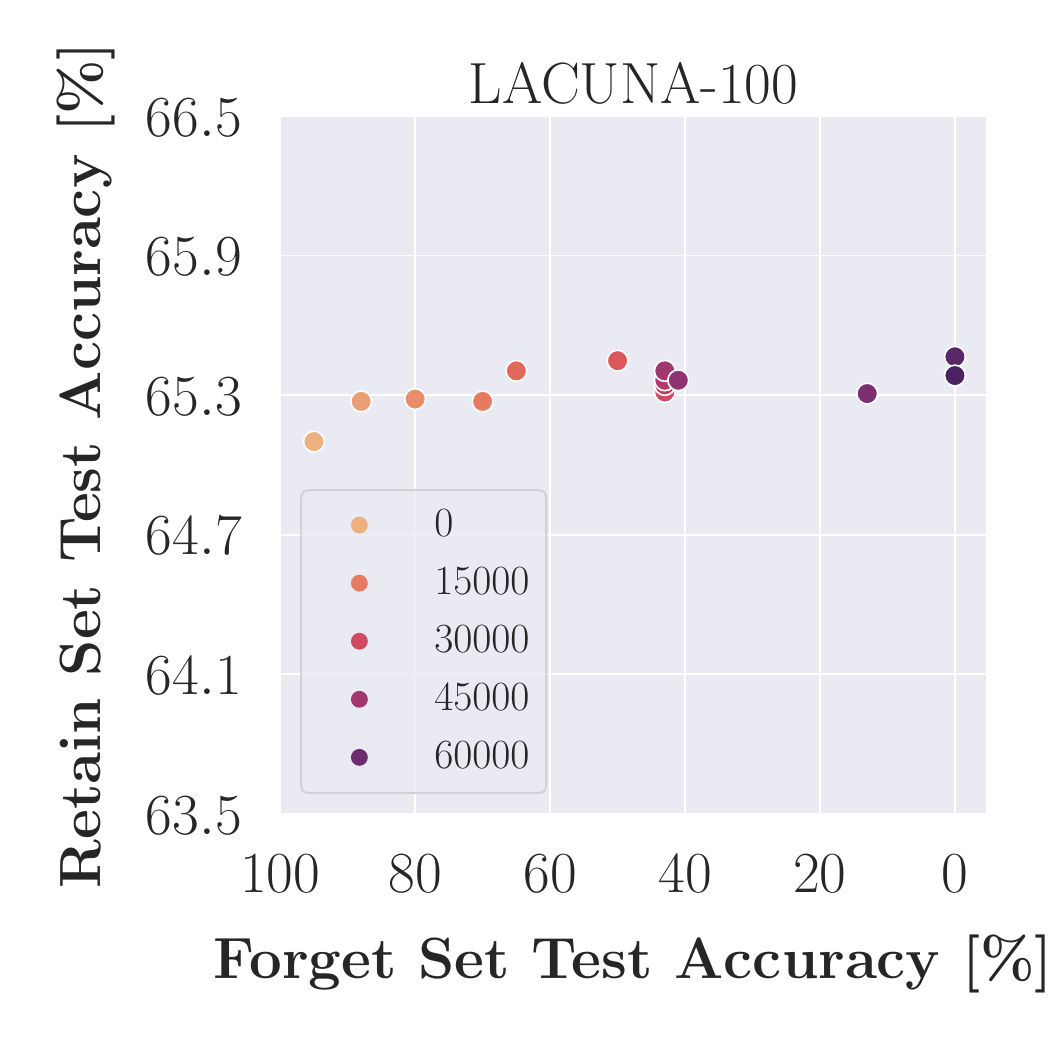}
    \includegraphics[width=0.23\linewidth]{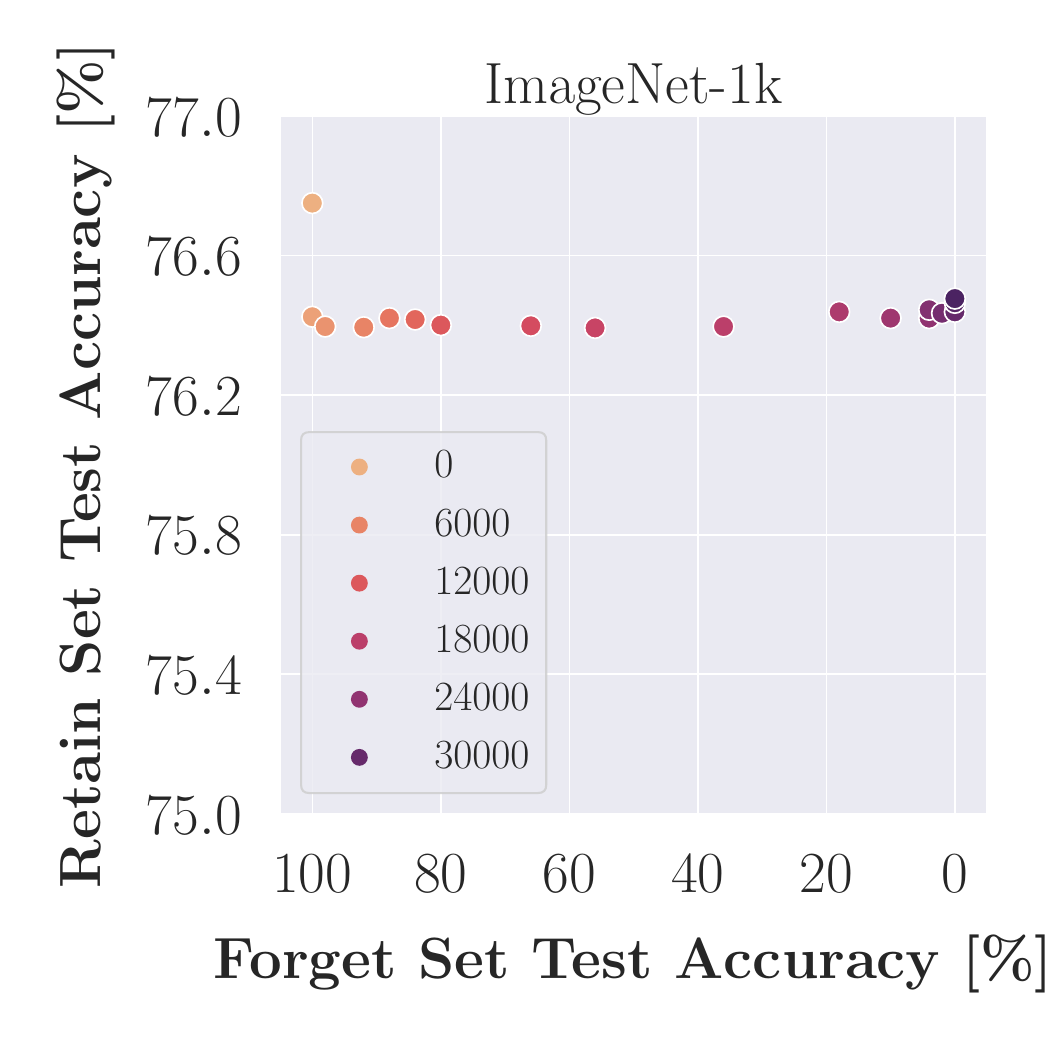}
    \label{fig2:b}
    \vspace{-0.2cm}
    \caption{\textbf{Unlearning via Activations.} Performance on the retain set test data vs. Performance on the forget set test data across various datasets for (a) CLIP pretrained ViT/B-32 in the \textbf{top row} (b) ImageNet pretrained ResNet-50 backbones in the \textbf{bottom row} as the value of $N_a$ is increased which is indicated by the color of the markers. The relative performance on the retain set test data as compared to the original models increases after unlearning in the case of CIFAR-10 and ImageNet-1k and drops for CIFAR-100 and LACUNA-100 in the case of ViT/B-32 and increases for all four datasets in the case of ResNet-50 (see Table 
    \ref{tab:baseline_compare}).}
    \vspace{-0.2cm}
    \label{fig2}
\end{figure*}

\textbf{Unlearning}
We aim to make the problem of unlearning as challenging as possible in order to fairly evaluate the proposed methods. Therefore, on each dataset we select the class that is best learned by the respective models with the Discrete Key Value Bottleneck trained previously, to be the forget class (see  Appendix~\ref{app:forget} for further details).


\textbf{Objective \& Metrics}
\label{sec:obj_metrics}
We report our results on the test data of retain classes and forget class, i.e.\ $\mathcal{D}_{test}^{retain}$ and $\mathcal{D}_{test}^{forget}$. Further, in our experiments, we aim to achieve \textit{complete unlearning} - achieving minimal accuracy on the forget set while incurring minimal damage to the performance on the retain set. While achieving complete unlearning may not always be desirable, such as in the case of Membership Inference Attacks (MIAs) the proposed methods can be easily extended to defend against MIAs (We refer to Appendix~\ref{app:mia} for further discussion on MIAs and the proposed methods). We report mean values across 5 random seeds in all cases. 

For comparing the compute efficiency of different approaches, we report the approximate FLOPs (Floating Point Operations) required for the procedure of unlearning. The total number of FLOPs are calculated as \textit{number of FLOPs required during the forward passes + number of FLOPs required during the backward passes}. We use the \texttt{fvcore}\footnote{https://github.com/facebookresearch/fvcore} library for computing the number FLOPs required during the forward passes. FLOPs required using backward passes are approximated as \textit{number of operations used for gradient computations + number of operations used for weight updates}\footnote{https://epochai.org/blog/backward-forward-FLOP-ratio}. Since only the linear head weights are trainable, the number of computations required for calculating the gradients would be the same as the number of parameters in the linear layer. Further, since we use Adam optimizer, the number of operations required for the weight updates would be equal to 18 times the number of parameters.

To calculate the final number of FLOPs, we first calculate the FLOPs required for one example (one forward + backward pass) and then multiply them with the total number of examples and the total number of epochs. For SCRUB, the forward and backwards FLOPs are multiplied with different scalars depending on the \textit{msteps} parameter.


\subsection{Unlearning via the Discrete Key-Value Bottleneck}
\label{subec:results_zeroshot}

We will now discuss the results of unlearning via activations and examples, i.e.\ the two approaches proposed in Section~\ref{sec:method} on all four benchmark datasets. 

\begin{figure*}[t]
    \centering
    \includegraphics[width=0.23\linewidth]{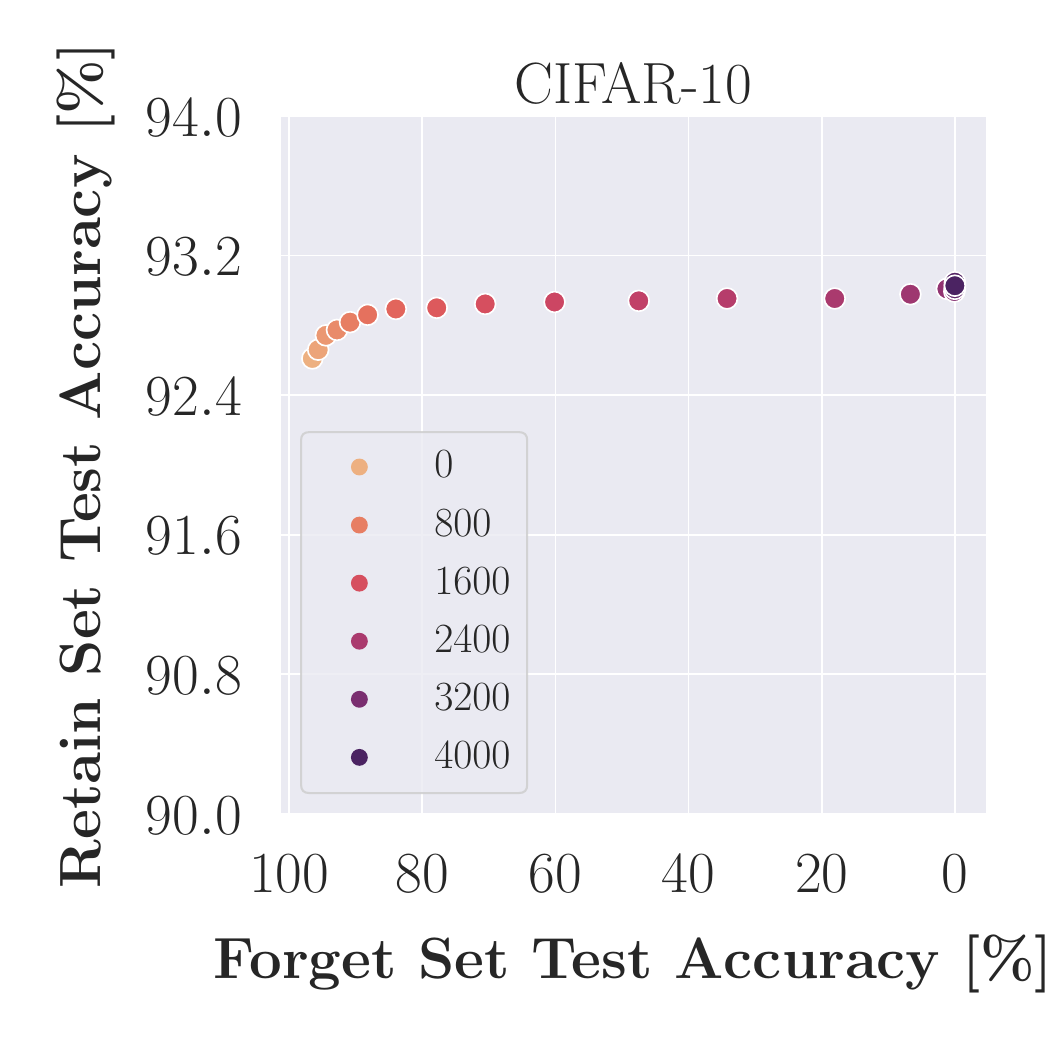}
    \includegraphics[width=0.23\linewidth]{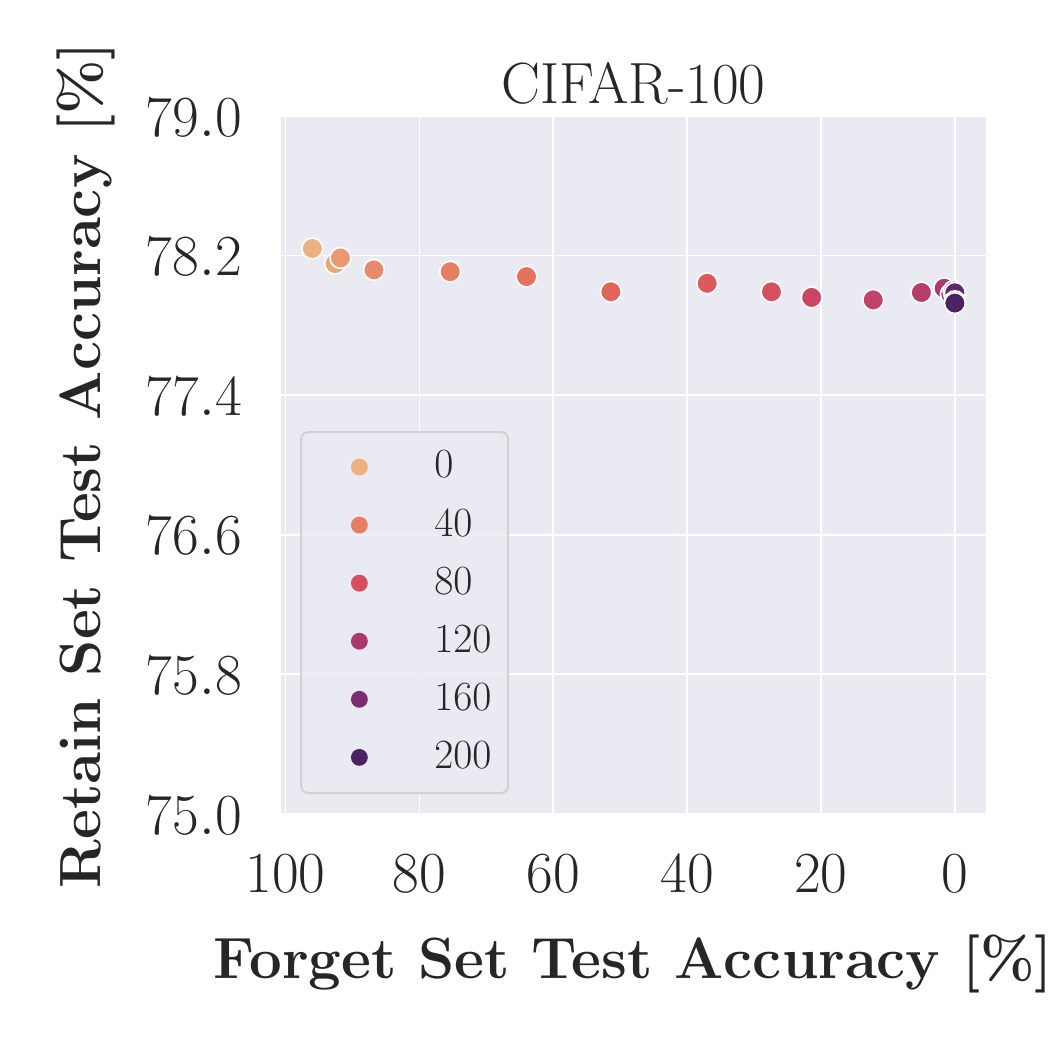}
    \includegraphics[width=0.23\linewidth]{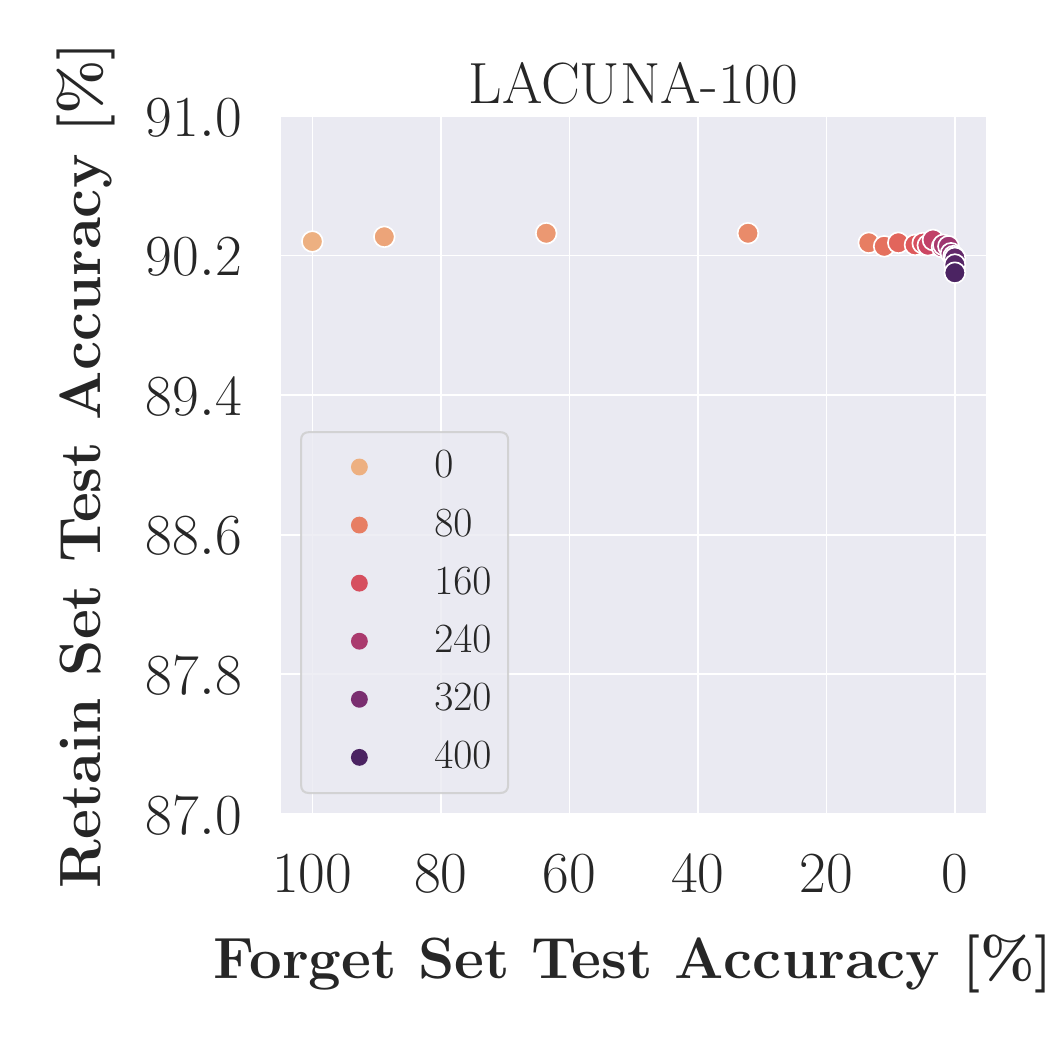}
    \includegraphics[width=0.23\linewidth]{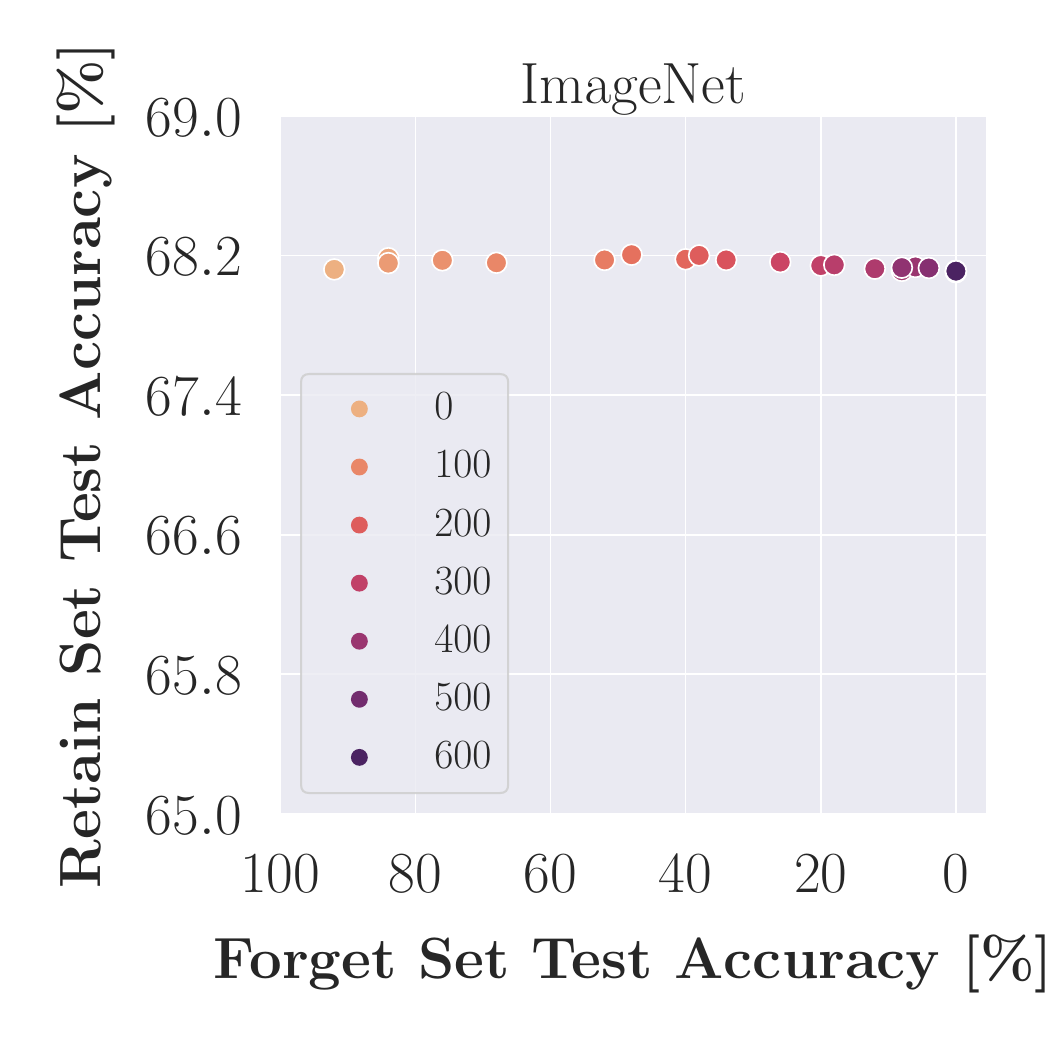}
    \label{fig3:a}
    \\
    \includegraphics[width=0.23\linewidth]{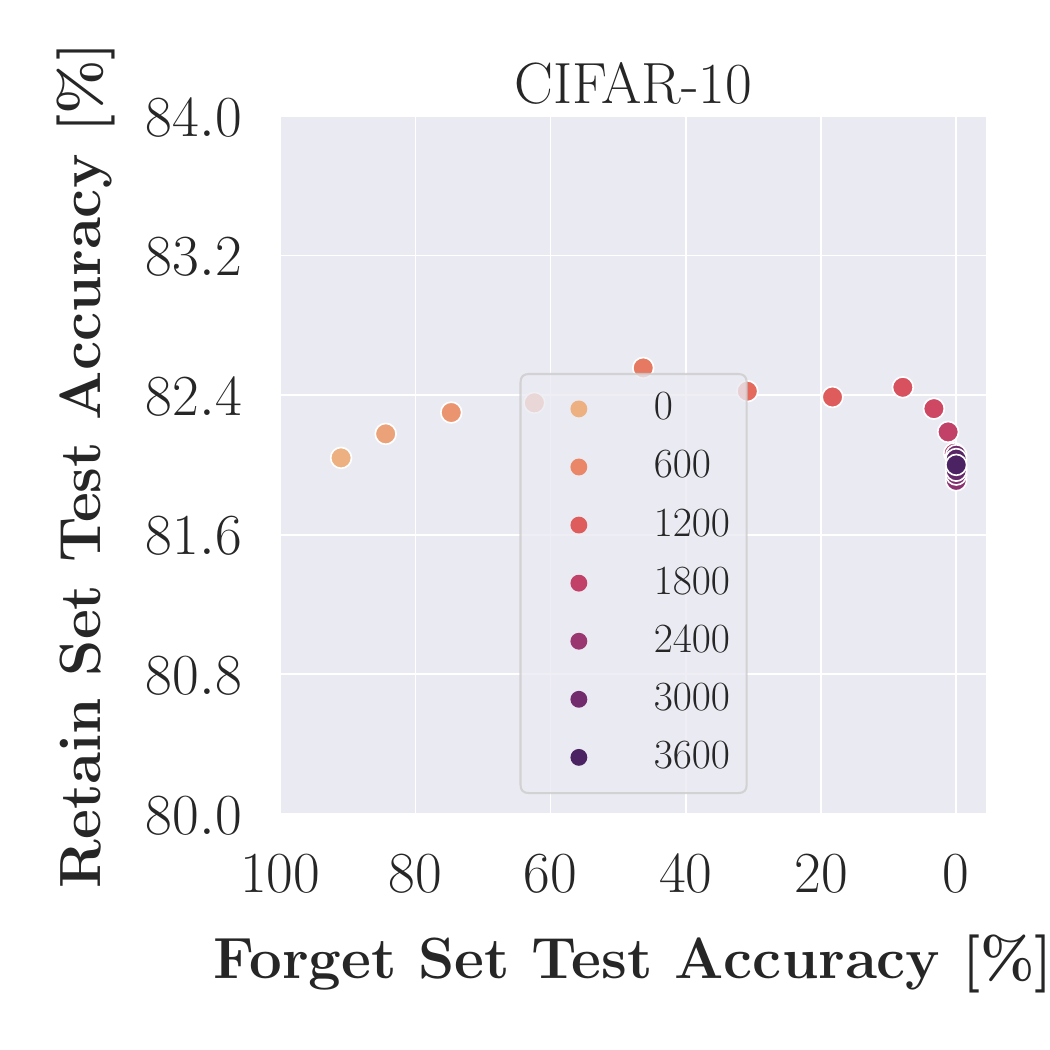}
    \includegraphics[width=0.23\linewidth]{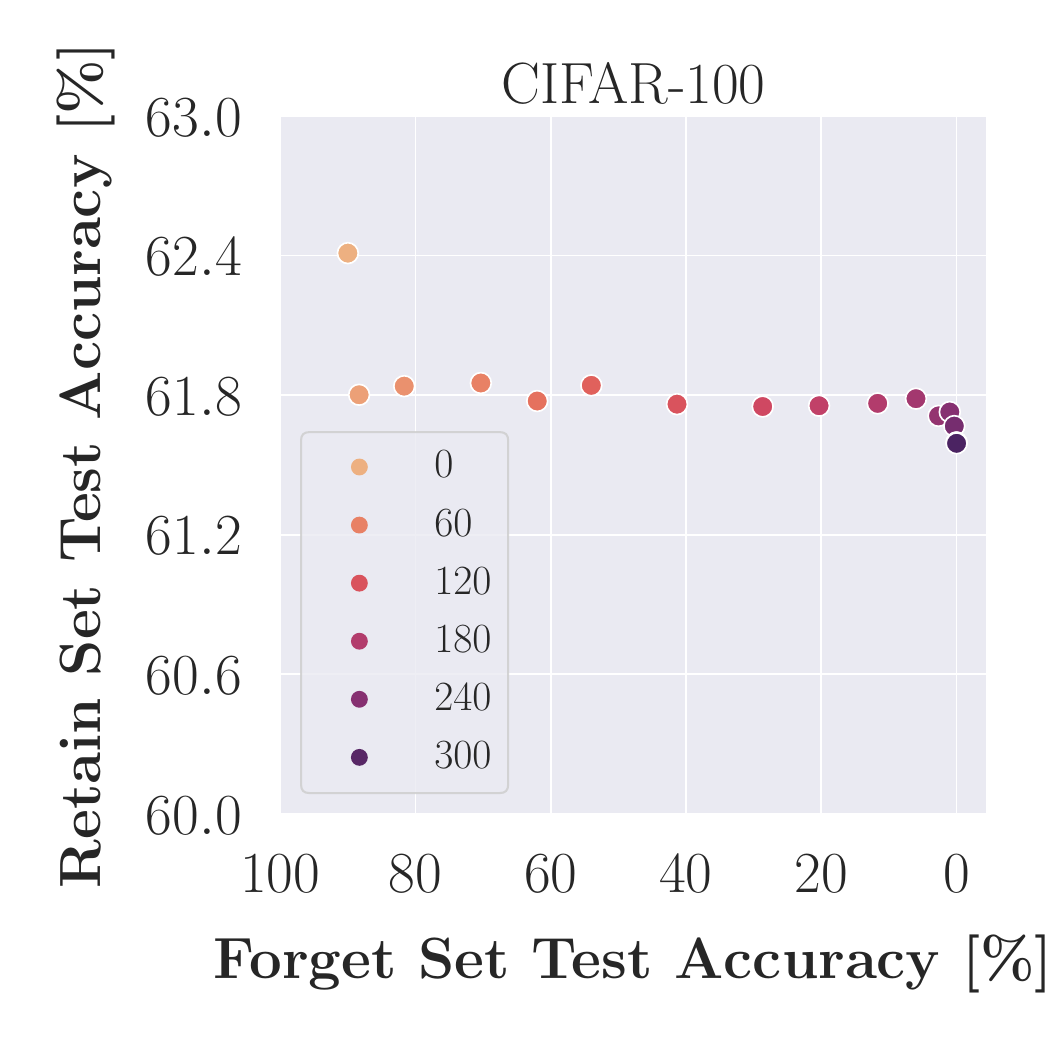}
    \includegraphics[width=0.23\linewidth]{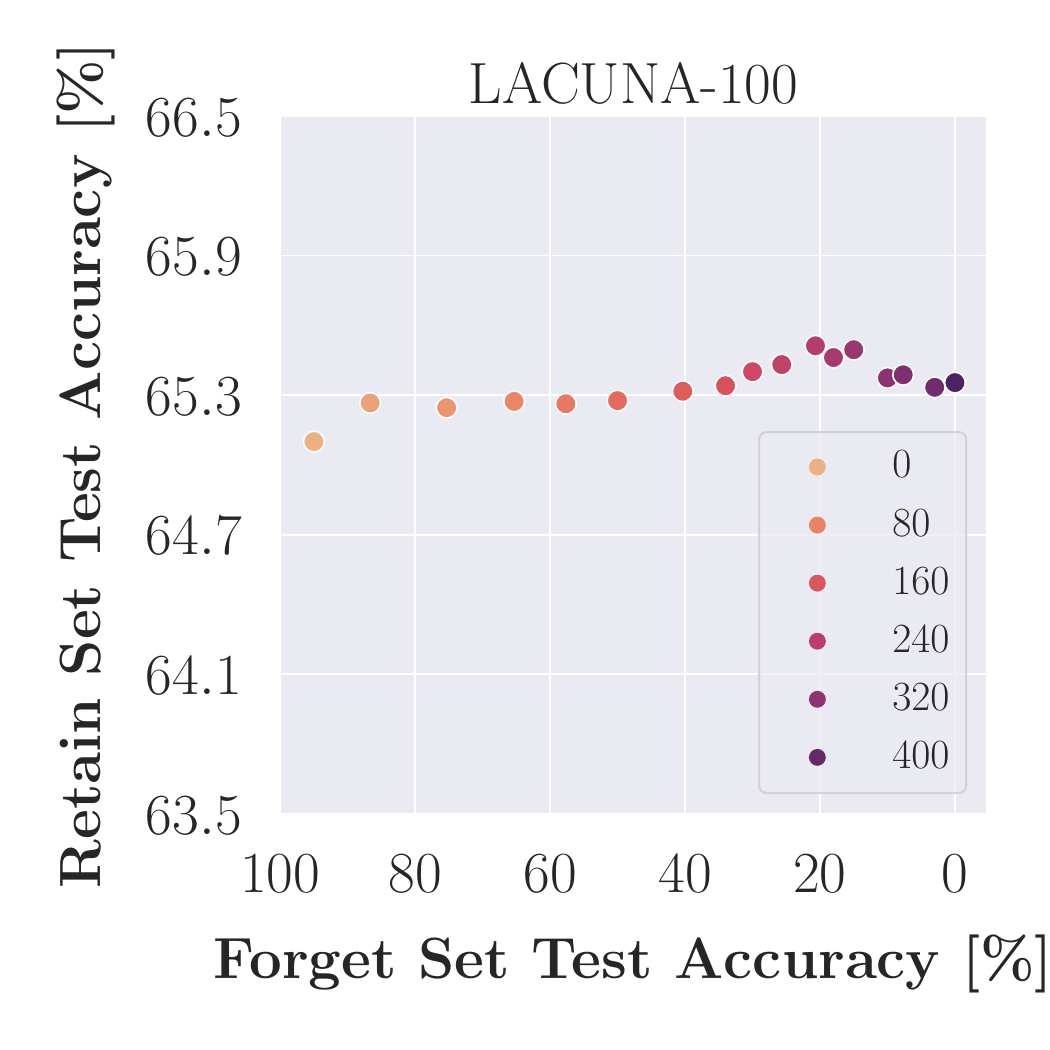}
    \includegraphics[width=0.23\linewidth]{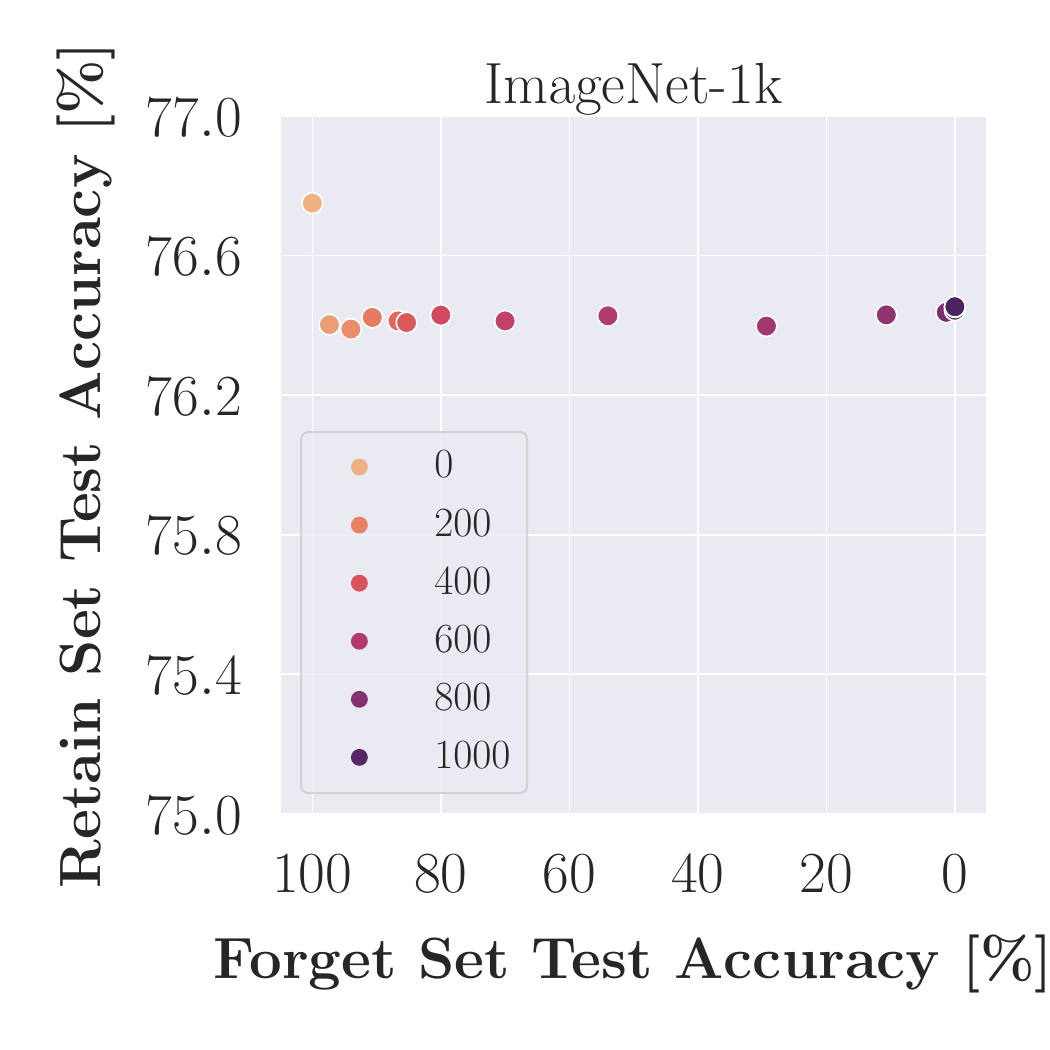}
    \label{fig3:b}
    \vspace{-0.2cm}
    \caption{\textbf{Unlearning via Examples}. Performance on the retain set test data vs. Performance on the forget set test data across different datasets for (a) CLIP pretrained ViT/B-32 in the \textbf{top row} and (b) ImageNet pretrained ResNet-50 backbones in the \textbf{bottom row} as the value of $N_e$ is increased which is indicated by the color of the markers. The relative performance on the retain set test data as compared to the base model increases for CIFAR-10 and drops for all other datasets in the case of ViT/B-32, whereas it drops for CIFAR-10 and CIFAR-100 and increases for LACUNA-100 and ImageNet-1k in the case of ResNet-50 (see Table \ref{tab:baseline_compare})}
    \vspace{-0.2cm}
    \label{fig3}
\end{figure*}

\textbf{Unlearning via Activations.}
Unlearning via activations requires us to set the hyperparameter $N_a$, reflecting the top-$N_a$ most frequently activated key-value pairs which will be masked out 
after inference on the forget set. We therefore start by analyzing its role over a wide range of values to probe its choice and effect with $N_a=0$ being the limit without any unlearning. Figure \ref{fig2} summarizes the unlearning and effect of $N_a$ on the retain vs forget test set. 
In the case of CIFAR-10 and a ViT/B-32 backbone, the initial accuracies on the retain and forget test set are 92.61\% and 96.50\% respectively. As $N_a$ increases, the forget class test accuracy decreases, slowly for small $N_a$ and rapidly for larger $N_a$. The model reaches random accuracy (i.e. 10\% for CIFAR-10) on the forget class test data at $N_a = 150000$. At this point the retain set test accuracy is 92.97\%. The model unlearns the forget class completely between $N_a = 170,000$ (0.4\%) and $N_a = 180,000$ (0\%). At this point, the retain set test accuracy is 92.94\%, which is almost identical to the initial accuracy. Further increasing $N_a$ up to $N_a = 200,000$, i.e. about 20\% of all key-value pairs, leads to an additional increase in retain set test accuracy to 93\%. On the contrary, in the case of CIFAR-10 and a ResNet-50 backbone, the decrease in the forget set test accuracy is rapid for small $N_a$ and it slow for higher values of $N_a$. Complete unlearning in this case happens between $N_a$ = 160000 (0.1\%) and $N_a$ = 190000 (0\%). These differences in trends can be attributed to how the information is factorized among the representations. For eg. there is a steep decline in the forget set test accuracy between $N_a$ = 50000 (41\%) and $N_a$ = 55000 (13\%) in the case of LACUNA-100 and ResNet-50 backbone (Figure~\ref{fig2:b}). This behavior may be attributed to the presence of high information but less frequently selected key-value pairs between the two values of $N_a$.
Nevertheless, as can be seen from the equivalent analysis on the CIFAR-100, LACUNA-100 and ImageNet-1k models in Figure~\ref{fig2}, the same trend of maintaining the initial retain accuracy while minimizing the forget accuracy up to a minimum holds across all four datasets and both backbones validating its meaningful unlearning capability.

\textbf{Unlearning via Examples.} For the second 
method---unlearning via examples---$N_e$ examples are sampled randomly from the training data of the forget class, and subsequently used for unlearning by the mechanism described in Section \ref{sec:method}. Similar to before, we aim to assess the effect on the choice of $N_e$ over a wide range for each dataset, including $N_e=0$ being the limit without any unlearning. Figure \ref{fig3} summarizes the unlearning and effect of $N_e$ on the retain vs. forget test set. 
We again begin by focusing on the results with CIFAR-10 and ViT/B-32 backbone. Here, the  forget set $\mathcal{D}_{train}^{forget}$ contains 5000 examples. 
We start off with retain set and forget set test accuracies of 92.61\% and 96.50\% respectively. Similar to the previous approach -- unlearning via activations -- the test accuracy on the forget set decreases with increasing $N_e$. The accuracy on the retain test set, on the other hand, increases monotonically, although only slightly overall. The model achieves random accuracy on the forget class around $N_e=2500$. The accuracy on retain set test data is at just under 93\% at this transition. Finally, the accuracy on the forget set drops to 0\% (i.e.\ complete unlearning) between $N_e = 3000$ and $N_e = 3400$ with a retain set test accuracy of just above 93\% at $N_e = 3400$. Further increasing $N_e$ does not affect the retain set test accuracy notably. Similarly to the case of unlearning via activations, the forget set test performance decreases rapidly at first and then slowly with $N_e$ in the case of CIFAR-10 with a ResNet-50 backbone. The retain set test accuracy increases at first and then decreases, albeit marginally. An equivalent analysis on the CIFAR-100, Lacuna-100 and ImageNet-1k models in Figure~\ref{fig3} exhibits a similar behavior of  successful minimization of the forget accuracy up to a minimum while roughly maintaining the retain set test accuracy, validating unlearning via examples as another option for unlearning using discrete key-value bottlenecks.


\textbf{Summary.} Both methods, \textit{Unlearning via Activations} and \textit{Unlearning via Examples}, successfully demonstrate unlearning of the forget class while having a negligible effect on the models' performance on the retain set. Importantly, this is achieved without any form of training, retraining, or fine-tuning as is usually required by other methods. The retain set test accuracy remains more or less constant for all four datasets except for a few minor fluctuations. This is a result of the fact that due to localized and context-dependent \textit{sparse updates} during the initial training of the model, discrete key-representations corresponding to different classes in the dataset are well separated from each other, an important prerequisite discussed in  \citet{trauble2023discrete}. 
Hence, all the information about a class can be unlearned by forgetting only a subset of the forget class training data in the case of \textit{Unlearning via Examples}, making it very data-efficient. 

\begin{table*}[t]
\caption{Comparison between the proposed methods and the baseline across CIFAR-10, CIFAR-100, LACUNA-100 and ImageNet-1k datasets and CLIP pretrained ViT/B-32 and ImageNet pretrained ResNet-50 backbones. We compare the \textbf{relative change in performance} on the retain and forget set test data relative to the originally trained models. The proposed methods are able to unlearn the forget sets completely in all cases while causing minimal changes in the performance of the models on the retain set test data.}
\vspace{-0.3cm}
\label{tab:baseline_compare}
\begin{center}
\resizebox{\textwidth}{!}{
\begin{tabular}{c|c|cc|cc|cc|cc}
\toprule
                                    & & \multicolumn{2}{c}{CIFAR-10}   & \multicolumn{2}{c}{CIFAR-100} & \multicolumn{2}{c}{LACUNA-100} & \multicolumn{2}{c}{ImageNet-1k} \\
\midrule
                            Backbone & Method & $\mathcal{D}_{test}^{retain}$   & $\mathcal{D}_{test}^{forget}$  & $\mathcal{D}_{test}^{retain}$ & $\mathcal{D}_{test}^{forget}$ & $\mathcal{D}_{test}^{retain}$ & $\mathcal{D}_{test}^{forget}$ & $\mathcal{D}_{test}^{retain}$ & $\mathcal{D}_{test}^{forget}$ \\
\midrule
\multirow{3}{*}{ViT/B-32} & DKVB via Activations (sec \ref{subec:results_zeroshot})  &  \textbf{0.36\%}     &-100\%       & -\textbf{0.20}\%     & -100\%     & -0.17\%     & -100\%  & 0.15\% & -100\%           \\
&DKVB via Examples (sec \ref{subec:results_zeroshot})     &  0.45\%      &-100\%      & -0.36\%     & -100\%     & \textbf{-0.09\%}     & -100\%   & \textbf{-0.03\%} & -100\%        \\
&Linear Layer + SCRUB        &  1.62\%      & -100\%      & -0.91\%     & -100\%     & -1.10\%     & -100\%  & 7.31\% & -100\%          \\
& Linear Layer + Finetuning & 1.94\% & -100\% & -1.91\% & -98.33\% & -2.21\% & -100\% & 0.88\% & -100\% \\
& Linear Layer + Retraining & 1.82\% & -100\% & -0.39\%  & -100\%& -2.03\% & -100\% & 5.16\% & -100\% \\
& Linear Layer + NegGrad+ & 0.49\% & -100\% &-0.63\%  &-100\% & -1.34\% & -100\% & 2.45\% & -100\% \\
\midrule
\multirow{3}{*}{ResNet-50} & DKVB via Activations (sec \ref{subec:results_zeroshot})  &    \textbf{0.04\%}   &   -100\%    &   \textbf{0.26\%}   &   -100\%   &  0.21\%   & -100\%  & \textbf{0.04\%} &    -100\%       \\
&DKVB via Examples (sec \ref{subec:results_zeroshot})     &    -0.07\%   &   -100\%   &   -0.34\%   &  -100\%    &  \textbf{0.17\%}    &  -100\%  & \textbf{0.04\%} & -100\%        \\
&Linear Layer + SCRUB        &    -0.07\%    &   -99.67\%    &  -0.94\%    &  -98.79\%    &   -0.26\%   & -99.67\%  & 0.74\% & -100\%          \\
& Linear Layer + Finetuning & 0.48\% &-100\% &-0.46\% &-99.99\% &-2.96\% & -100\% & -2.25\% & -100\% \\
& Linear Layer + Retraining & 3.06\% & -100\% & 1.76\% & -100\% & 1.15\% & -100\% & -1.14\% & -100\% \\
& Linear Layer + NegGrad+ & 2.13\% & -100\% & -0.85\% & -100\% &6.73\% & -100\% &-0.85\%  & -100\% \\
\bottomrule
\end{tabular}
}              
\end{center}
\vspace{-0.3cm}
\end{table*}

\subsubsection{Comparison with Baselines}
\label{sec:compbase}

We now compare the results of both the proposed methods, which require Backbone + DKVB models against several baseline methods, which are optimized for models without such a bottleneck. For this, we will use the Backbone + Linear Layer models described in \ref{subsec:exp_setup}. 
On these models, we run SCRUB~\citep{kurmanji2023towards}, \textit{finetuning} - finetuning the model to be unlearnt on the retain set, \textit{retraining} - training the model from scratch on the retain set only and NegGrad+~\citep{kurmanji2023towards} and compare the performance changes on the forget and retain classes against the performance changes after unlearning with the two proposed methods.
Table~\ref{tab:baseline_compare} shows the comparison between the two previously reported methods and the baselines.
We can see that one of the two proposed approaches always results in the least change in the performance of the base model on the retain classes, while at the same time achieving complete unlearning of the forget class. 
The baselines on the other hand, occasionally fail to achieve complete unlearning.
Finally, it is important to re-emphasize that the proposed methods achieve the shown performance without requiring any additional gradient-based training for unlearning. In the case of baselines, we stop the unlearning procedure when the forget set is completely unlearned or the forget set test accuracy has converged with minimal damage to the performance on the retain set. Moreover, while we report results for the case of complete unlearning, the proposed methods can be easily used for achieving unlearning of the forget class to different extents by tuning the $N_a$ and $N_e$ hyperparameters. We refer to Appendix~\ref{hp:scrub} and  \ref{hp:base} for further training and implementation details.\looseness=-1

\vspace{-2mm}
\subsection{Proposed Methods achieve Unlearning in a Compute Efficient Manner}
\label{sec:comp_eff}
\vspace{-2mm}
In this section, we compare the proposed approaches against the baselines in terms of the amount of compute required in order to achieve complete unlearning. To facilitate this comparison, we report the number of FLOPs required for the unlearning procedure for each case. The FLOPs are calculated following the rules described in Section~\ref{sec:obj_metrics}. Table~\ref{tab:flops_compare} compared the FLOPs required for unlearning in each case.

\begin{table*}[t]
\caption{Comparison of FLOPs for various methods across CIFAR-10, CIFAR-100, LACUNA-100, and ImageNet-1k datasets using ViT/B-32 and ResNet-50 backbones. We report both forward and backward FLOPs for each method.}
\vspace{-0.3cm}
\label{tab:flops_compare}
\begin{center}
\resizebox{\textwidth}{!}{
\begin{tabular}{c|c|cc|cc|cc|cc}
\toprule
 & & \multicolumn{2}{c}{CIFAR-10} & \multicolumn{2}{c}{CIFAR-100} & \multicolumn{2}{c}{LACUNA-100} & \multicolumn{2}{c}{ImageNet-1k} \\
\midrule
Backbone & Method & Forward & Backward & Forward & Backward & Forward & Backward & Forward & Backward \\
& & (TFLOPs) & (GFLOPs) & (TFLOPs) & (GFLOPs) & (TFLOPs) & (GFLOPs) & (TFLOPs) & (GFLOPs) \\
\midrule
\multirow{6}{*}{ViT/B-32} 
& DKVB via Activations (sec~\ref{subec:results_zeroshot})          & 21.93 & \textbf{0}       & 2.19  & \textbf{0}       & 1.75 & \textbf{0}       & 5.63  & \textbf{0}         \\
& DKVB via Examples (sec~\ref{subec:results_zeroshot})          & \textbf{14.91}         & \textbf{0}       & \textbf{0.75}           & \textbf{0}       & \textbf{1.40}          & \textbf{0}       & \textbf{2.90}           & \textbf{0}         \\
& Linear Layer + SCRUB        & 655.13         & 14.59  & 1316.83        & 293.30 & 527.60        & 117.51 & 39168.28       & 87230.98 \\
& Linear Layer + Finetuning   & 196.54         & 4.38  & 6485.87        & 1444.61 & 5188.69       & 1155.69 & 11179.75      & 24898.21 \\
& Linear Layer + Retraining   & 196.54         & 4.38  & 1080.98        & 240.77 & 864.78        & 192.61& 5589.87        & 12449.11 \\
& Linear Layer + NegGrad+     & 393.08         & 8.76  & 1729.56        & 385.23 & 1383.65       & 308.18& 11179.75       & 24898.21 \\
\midrule
\multirow{6}{*}{ResNet-50} 
& DKVB via Activations (sec~\ref{subec:results_zeroshot})          & 16.66 & \textbf{0}       & 1.67  & \textbf{0}       & 1.33 & \textbf{0}       & 5.31  & \textbf{0}         \\
& DKVB via Examples (sec~\ref{subec:results_zeroshot})          & \textbf{7.33}           & \textbf{0}       & \textbf{0.93}          & \textbf{0}       & \textbf{1.33}          & \textbf{0}       & \textbf{3.74}           & \textbf{0}         \\
& Linear Layer + SCRUB        & 1498.74        & 87.55  & 1488.79        & 869.68 & 3697.00       & 2159.62 & 31873.28       & 299227.17 \\
& Linear Layer + Finetuning   & 1049.12        & 61.29 & 1648.66        & 963.07 & 131.89        & 77.05 & 5304.25        & 49796.43 \\
& Linear Layer + Retraining   & 659.47         & 385.23 & 1648.66        & 963.07 & 2242.18       & 1309.78 & 5304.25       & 49796.43 \\
& Linear Layer + NegGrad+     & 329.73         & 192.61 & 10608.49       & 99592.85 & 263.79     & 154.09 & 5304.25       & 49796.43 \\
\bottomrule
\end{tabular}
}
\end{center}
\vspace{-0.3cm}
\end{table*}

We report the forward and backward FLOPs separately to highlight that the proposed approaches do not require any gradient based updates. Additionally, while the scale of backward FLOPs may seem insignificant against the forward FLOPs, it can easily blow up in cases where complex parametric decoders are used on top of the DKVB. In our experiments, the decoder is simply an average pooling layer. Nevertheless, we can see that the proposed approaches require significantly less forward FLOPs as compared to the baselines. This can be explained by the fact that the proposed approaches require only one forward pass through the models per example of the forget class training data, in order to cache the activations. The baseline methods on the other hand, require multiple forward passes, each corresponding to a single training epoch. 

\vspace{-2mm}
\section{Limitations and Future Work}
\vspace{-2mm}

The proposed methods inherit the limitations of the DKVB \citep{trauble2023discrete} such as  the reliance of DKVB  on pre-trained encoders which can extract meaningful shared representations and trade-offs in downstream performance due to the use of an information bottleneck. Extensions to the model may involve training sparse representations inducing discrete bottleneck end-to-end. Further, in our experiments, we consider the setting of \textit{class incremental} learning where the forget set can be easily identified and isolated. However, this may not always be sufficint for a given task and more complicated approaches might be needed to identify the data that needs to be removed from the model. 
While the methods introduced in this work are currently not designed for selective unlearning as outlined in Appendix \ref{app:mia}, there are various directions for future adaptations. These directions include enhancing the forget set isolation, and addressing limitations related to information retention, for instance by further fine-tuning on the encoder backbone. Advancing towards even stronger unlearning forms involving discrete key value bottlenecks, represents a key direction for future work.

\vspace{-2mm}
\section{Conclusion}
\vspace{-2mm}
In this work, we proposed a new approach to unlearning that requires minimal computation in order to unlearn a subset of data. This approach is based on the use of a discrete architectural bottleneck which induces sparse representations. These sparse representations facilitate unlearning a subset of data from the model with minimal to no performance drop on the rest of the data. We focused on the setting of class unlearning and our experiments show that the proposed approach, while being at least 20$\times$ compute efficient, performs competitively with or in some cases better than a state-of-the-art approach which requires additional compute to perform unlearning. Consequently, excising the activated key-value pairs from the model is a highly effective means of unlearning the forget set without disrupting the retain set.

\section{Acknowledgements}
This research was enabled by compute resources provided by Mila (mila.quebec). VS would like to thank Aniket Didolkar and Moksh Jain for useful discussions, feedback on the initial versions of the paper and proof reading the paper. VS would also like to thank Nanda H Krishna and Moksh Jain for helping with the figures included in the paper and Thomas Jiralerspong for proof reading and giving feedback on the paper.

\bibliography{iclr2025_conference}

\begin{thebibliography}{41}
\providecommand{\natexlab}[1]{#1}
\providecommand{\url}[1]{\texttt{#1}}
\expandafter\ifx\csname urlstyle\endcsname\relax
  \providecommand{\doi}[1]{doi: #1}\else
  \providecommand{\doi}{doi: \begingroup \urlstyle{rm}\Url}\fi

\bibitem[Cao et~al.(2018)Cao, Shen, Xie, Parkhi, and Zisserman]{Cao18}
Qiong Cao, Li~Shen, Weidi Xie, Omkar~M. Parkhi, and Andrew Zisserman.
\newblock {VGGFace2}: A dataset for recognising faces across pose and age.
\newblock In \emph{International Conference on Automatic Face and Gesture
  Recognition}, 2018.

\bibitem[Cao \& Yang(2015)Cao and Yang]{7163042}
Yinzhi Cao and Junfeng Yang.
\newblock Towards making systems forget with machine unlearning.
\newblock In \emph{2015 IEEE Symposium on Security and Privacy}, pp.\
  463--480, 2015.
\newblock \doi{10.1109/SP.2015.35}.

\bibitem[Cauwenberghs \& Poggio(2000)Cauwenberghs and
  Poggio]{cauwenberghs2000incremental}
Gert Cauwenberghs and Tomaso Poggio.
\newblock Incremental and decremental support vector machine learning.
\newblock \emph{Advances in neural information processing systems}, 13, 2000.

\bibitem[Chen et~al.(2023)Chen, Gao, Liu, Peng, and Wang]{Chen2023BoundaryUR}
Min Chen, Weizhuo Gao, Gaoyang Liu, Kai Peng, and Chen Wang.
\newblock Boundary unlearning: Rapid forgetting of deep networks via shifting
  the decision boundary.
\newblock \emph{2023 IEEE/CVF Conference on Computer Vision and Pattern
  Recognition (CVPR)}, pp.\  7766--7775, 2023.
\newblock URL \url{https://api.semanticscholar.org/CorpusID:257636742}.

\bibitem[Chundawat et~al.(2023)Chundawat, Tarun, Mandal, and
  Kankanhalli]{chundawat2023zero}
Vikram~S Chundawat, Ayush~K Tarun, Murari Mandal, and Mohan Kankanhalli.
\newblock Zero-shot machine unlearning.
\newblock \emph{IEEE Transactions on Information Forensics and Security}, 2023.

\bibitem[Dang(2021)]{10.1007/978-3-030-71782-7_35}
Quang-Vinh Dang.
\newblock Right to be forgotten in the age of machine learning.
\newblock In Tatiana Antipova (ed.), \emph{Advances in Digital Science}, pp.\
  403--411, Cham, 2021. Springer International Publishing.
\newblock ISBN 978-3-030-71782-7.

\bibitem[Dosovitskiy et~al.(2020)Dosovitskiy, Beyer, Kolesnikov, Weissenborn,
  Zhai, Unterthiner, Dehghani, Minderer, Heigold, Gelly,
  et~al.]{dosovitskiy2020image}
Alexey Dosovitskiy, Lucas Beyer, Alexander Kolesnikov, Dirk Weissenborn,
  Xiaohua Zhai, Thomas Unterthiner, Mostafa Dehghani, Matthias Minderer, Georg
  Heigold, Sylvain Gelly, et~al.
\newblock An image is worth 16x16 words: Transformers for image recognition at
  scale.
\newblock \emph{arXiv preprint arXiv:2010.11929}, 2020.

\bibitem[Duan et~al.(2007)Duan, Li, He, and Zeng]{Duan2007DecrementalLA}
Hua Duan, Huan Li, Guoping He, and Qingtian Zeng.
\newblock Decremental learning algorithms for nonlinear langrangian and least
  squares support vector machines.
\newblock 2007.
\newblock URL \url{https://api.semanticscholar.org/CorpusID:13986244}.

\bibitem[Esser et~al.(2021)Esser, Rombach, and Ommer]{esser2021taming}
Patrick Esser, Robin Rombach, and Bjorn Ommer.
\newblock Taming transformers for high-resolution image synthesis.
\newblock In \emph{Proceedings of the IEEE/CVF conference on computer vision
  and pattern recognition}, pp.\  12873--12883, 2021.

\bibitem[Frankle \& Carbin(2018)Frankle and Carbin]{frankle2018lottery}
Jonathan Frankle and Michael Carbin.
\newblock The lottery ticket hypothesis: Finding sparse, trainable neural
  networks.
\newblock \emph{arXiv preprint arXiv:1803.03635}, 2018.

\bibitem[Ginart et~al.(2019)Ginart, Guan, Valiant, and Zou]{ginart2019making}
Antonio Ginart, Melody Guan, Gregory Valiant, and James~Y Zou.
\newblock Making ai forget you: Data deletion in machine learning.
\newblock \emph{Advances in neural information processing systems}, 32, 2019.

\bibitem[Golatkar et~al.(2020{\natexlab{a}})Golatkar, Achille, and
  Soatto]{golatkar2020eternal}
Aditya Golatkar, Alessandro Achille, and Stefano Soatto.
\newblock Eternal sunshine of the spotless net: Selective forgetting in deep
  networks.
\newblock In \emph{Proceedings of the IEEE/CVF Conference on Computer Vision
  and Pattern Recognition}, pp.\  9304--9312, 2020{\natexlab{a}}.

\bibitem[Golatkar et~al.(2020{\natexlab{b}})Golatkar, Achille, and
  Soatto]{golatkar2020forgetting}
Aditya Golatkar, Alessandro Achille, and Stefano Soatto.
\newblock Forgetting outside the box: Scrubbing deep networks of information
  accessible from input-output observations.
\newblock In \emph{Computer Vision--ECCV 2020: 16th European Conference,
  Glasgow, UK, August 23--28, 2020, Proceedings, Part XXIX 16}, pp.\  383--398.
  Springer, 2020{\natexlab{b}}.

\bibitem[Goyal et~al.(2021)Goyal, Didolkar, Lamb, Badola, Ke, Rahaman, Binas,
  Blundell, Mozer, and Bengio]{goyal2021coordination}
Anirudh Goyal, Aniket Didolkar, Alex Lamb, Kartikeya Badola, Nan~Rosemary Ke,
  Nasim Rahaman, Jonathan Binas, Charles Blundell, Michael Mozer, and Yoshua
  Bengio.
\newblock Coordination among neural modules through a shared global workspace.
\newblock \emph{arXiv preprint arXiv:2103.01197}, 2021.

\bibitem[Guo et~al.(2019)Guo, Goldstein, Hannun, and Van
  Der~Maaten]{guo2019certified}
Chuan Guo, Tom Goldstein, Awni Hannun, and Laurens Van Der~Maaten.
\newblock Certified data removal from machine learning models.
\newblock \emph{arXiv preprint arXiv:1911.03030}, 2019.

\bibitem[Izzo et~al.(2021)Izzo, Smart, Chaudhuri, and Zou]{izzo2021approximate}
Zachary Izzo, Mary~Anne Smart, Kamalika Chaudhuri, and James Zou.
\newblock Approximate data deletion from machine learning models.
\newblock In \emph{International Conference on Artificial Intelligence and
  Statistics}, pp.\  2008--2016. PMLR, 2021.

\bibitem[Jacot et~al.(2018)Jacot, Gabriel, and Hongler]{jacot2018neural}
Arthur Jacot, Franck Gabriel, and Cl{\'e}ment Hongler.
\newblock Neural tangent kernel: Convergence and generalization in neural
  networks.
\newblock \emph{Advances in neural information processing systems}, 31, 2018.

\bibitem[Jaegle et~al.(2021)Jaegle, Gimeno, Brock, Vinyals, Zisserman, and
  Carreira]{jaegle2021perceiver}
Andrew Jaegle, Felix Gimeno, Andy Brock, Oriol Vinyals, Andrew Zisserman, and
  Joao Carreira.
\newblock Perceiver: General perception with iterative attention.
\newblock In \emph{International conference on machine learning}, pp.\
  4651--4664. PMLR, 2021.

\bibitem[Jia et~al.(2023)Jia, Liu, Ram, Yao, Liu, Liu, Sharma, and
  Liu]{jia2023model}
Jinghan Jia, Jiancheng Liu, Parikshit Ram, Yuguang Yao, Gaowen Liu, Yang Liu,
  Pranay Sharma, and Sijia Liu.
\newblock Model sparsification can simplify machine unlearning.
\newblock \emph{arXiv preprint arXiv:2304.04934}, 2023.

\bibitem[Krizhevsky et~al.(2009)Krizhevsky, Hinton,
  et~al.]{krizhevsky2009learning}
Alex Krizhevsky, Geoffrey Hinton, et~al.
\newblock Learning multiple layers of features from tiny images.
\newblock 2009.

\bibitem[Kurmanji et~al.(2023)Kurmanji, Triantafillou, and
  Triantafillou]{kurmanji2023towards}
Meghdad Kurmanji, Peter Triantafillou, and Eleni Triantafillou.
\newblock Towards unbounded machine unlearning.
\newblock \emph{arXiv preprint arXiv:2302.09880}, 2023.

\bibitem[Kwak et~al.(2017)Kwak, Lee, Park, and Lee]{Kwak2017LetMU}
Chanhee Kwak, Junyeong Lee, Kyuhong Park, and Heeseok Lee.
\newblock Let machines unlearn - machine unlearning and the right to be
  forgotten.
\newblock In \emph{Americas Conference on Information Systems}, 2017.
\newblock URL \url{https://api.semanticscholar.org/CorpusID:10605911}.

\bibitem[Liu et~al.(2021)Liu, Lamb, Kawaguchi, ALIAS PARTH~GOYAL, Sun, Mozer,
  and Bengio]{liu2021discrete}
Dianbo Liu, Alex~M Lamb, Kenji Kawaguchi, Anirudh~Goyal ALIAS PARTH~GOYAL, Chen
  Sun, Michael~C Mozer, and Yoshua Bengio.
\newblock Discrete-valued neural communication.
\newblock \emph{Advances in Neural Information Processing Systems},
  34:\penalty0 2109--2121, 2021.

\bibitem[Liu et~al.(2023)Liu, Shah, Boussif, Meo, Goyal, Shu, Mozer, Heess, and
  Bengio]{liu2023stateful}
Dianbo Liu, Vedant Shah, Oussama Boussif, Cristian Meo, Anirudh Goyal, Tianmin
  Shu, Michael~Curtis Mozer, Nicolas Heess, and Yoshua Bengio.
\newblock Stateful active facilitator: Coordination and environmental
  heterogeneity in cooperative multi-agent reinforcement learning.
\newblock In \emph{The Eleventh International Conference on Learning
  Representations}, 2023.
\newblock URL \url{https://openreview.net/forum?id=B4maZQLLW0_}.

\bibitem[Magdziarczyk(2019)]{inproceedings}
Małgorzata Magdziarczyk.
\newblock Right to be forgotten in light of regulation (eu) 2016/679 of the
  european parliament and of the council of 27 april 2016 on the protection of
  natural persons with regard to the processing of personal data and on the
  free movement of such data, and repealing directive 95/46/ec.
\newblock 04 2019.
\newblock \doi{10.5593/sgemsocial2019V/1.1/S02.022}.

\bibitem[Mantelero(2013)]{MANTELERO2013229}
Alessandro Mantelero.
\newblock The eu proposal for a general data protection regulation and the
  roots of the ‘right to be forgotten’.
\newblock \emph{Computer Law \& Security Review}, 29\penalty0 (3):\penalty0
  229--235, 2013.
\newblock ISSN 0267-3649.
\newblock \doi{https://doi.org/10.1016/j.clsr.2013.03.010}.
\newblock URL
  \url{https://www.sciencedirect.com/science/article/pii/S0267364913000654}.

\bibitem[Mehta et~al.(2022)Mehta, Pal, Singh, and Ravi]{mehta2022deep}
Ronak Mehta, Sourav Pal, Vikas Singh, and Sathya~N Ravi.
\newblock Deep unlearning via randomized conditionally independent hessians.
\newblock In \emph{Proceedings of the IEEE/CVF Conference on Computer Vision
  and Pattern Recognition}, pp.\  10422--10431, 2022.

\bibitem[Nguyen et~al.(2022)Nguyen, Huynh, Nguyen, Liew, Yin, and
  Nguyen]{nguyen2022survey}
Thanh~Tam Nguyen, Thanh~Trung Huynh, Phi~Le Nguyen, Alan Wee-Chung Liew,
  Hongzhi Yin, and Quoc Viet~Hung Nguyen.
\newblock A survey of machine unlearning.
\newblock \emph{arXiv preprint arXiv:2209.02299}, 2022.

\bibitem[Oord et~al.(2017)Oord, Vinyals, and Kavukcuoglu]{oord2017neural}
Aaron van~den Oord, Oriol Vinyals, and Koray Kavukcuoglu.
\newblock Neural discrete representation learning.
\newblock \emph{arXiv preprint arXiv:1711.00937}, 2017.

\bibitem[Pardau(2018)]{pardau2018california}
Stuart~L Pardau.
\newblock The california consumer privacy act: Towards a european-style privacy
  regime in the united states.
\newblock \emph{J. Tech. L. \& Pol'y}, 23:\penalty0 68, 2018.

\bibitem[Radford et~al.(2021)Radford, Kim, Hallacy, Ramesh, Goh, Agarwal,
  Sastry, Askell, Mishkin, Clark, et~al.]{radford2021learning}
Alec Radford, Jong~Wook Kim, Chris Hallacy, Aditya Ramesh, Gabriel Goh,
  Sandhini Agarwal, Girish Sastry, Amanda Askell, Pamela Mishkin, Jack Clark,
  et~al.
\newblock Learning transferable visual models from natural language
  supervision.
\newblock In \emph{International conference on machine learning}, pp.\
  8748--8763. PMLR, 2021.

\bibitem[Razavi et~al.(2019)Razavi, Van~den Oord, and
  Vinyals]{razavi2019generating}
Ali Razavi, Aaron Van~den Oord, and Oriol Vinyals.
\newblock Generating diverse high-fidelity images with vq-vae-2.
\newblock \emph{Advances in neural information processing systems}, 32, 2019.

\bibitem[Russakovsky et~al.(2015)Russakovsky, Deng, Su, Krause, Satheesh, Ma,
  Huang, Karpathy, Khosla, Bernstein, et~al.]{russakovsky2015imagenet}
Olga Russakovsky, Jia Deng, Hao Su, Jonathan Krause, Sanjeev Satheesh, Sean Ma,
  Zhiheng Huang, Andrej Karpathy, Aditya Khosla, Michael Bernstein, et~al.
\newblock Imagenet large scale visual recognition challenge.
\newblock \emph{International journal of computer vision}, 115:\penalty0
  211--252, 2015.

\bibitem[Shokri et~al.(2017)Shokri, Stronati, Song, and
  Shmatikov]{shokri2017membership}
Reza Shokri, Marco Stronati, Congzheng Song, and Vitaly Shmatikov.
\newblock Membership inference attacks against machine learning models.
\newblock In \emph{2017 IEEE symposium on security and privacy (SP)}, pp.\
  3--18. IEEE, 2017.

\bibitem[Tarun et~al.(2023)Tarun, Chundawat, Mandal, and
  Kankanhalli]{tarun2023fast}
Ayush~K Tarun, Vikram~S Chundawat, Murari Mandal, and Mohan Kankanhalli.
\newblock Fast yet effective machine unlearning.
\newblock \emph{IEEE Transactions on Neural Networks and Learning Systems},
  2023.

\bibitem[Tr{\"a}uble et~al.(2023)Tr{\"a}uble, Goyal, Rahaman, Mozer, Kawaguchi,
  Bengio, and Sch{\"o}lkopf]{trauble2023discrete}
Frederik Tr{\"a}uble, Anirudh Goyal, Nasim Rahaman, Michael~Curtis Mozer, Kenji
  Kawaguchi, Yoshua Bengio, and Bernhard Sch{\"o}lkopf.
\newblock Discrete key-value bottleneck.
\newblock In \emph{International Conference on Machine Learning}, pp.\
  34431--34455. PMLR, 2023.

\bibitem[Tsai et~al.(2014)Tsai, Lin, and Lin]{10.1145/2623330.2623661}
Cheng-Hao Tsai, Chieh-Yen Lin, and Chih-Jen Lin.
\newblock Incremental and decremental training for linear classification.
\newblock In \emph{Proceedings of the 20th ACM SIGKDD International Conference
  on Knowledge Discovery and Data Mining}, KDD '14, pp.\  343–352, New York,
  NY, USA, 2014. Association for Computing Machinery.
\newblock ISBN 9781450329569.
\newblock \doi{10.1145/2623330.2623661}.
\newblock URL \url{https://doi.org/10.1145/2623330.2623661}.

\bibitem[Villaronga et~al.(2018)Villaronga, Kieseberg, and
  Li]{VILLARONGA2018304}
Eduard~Fosch Villaronga, Peter Kieseberg, and Tiffany Li.
\newblock Humans forget, machines remember: Artificial intelligence and the
  right to be forgotten.
\newblock \emph{Computer Law \& Security Review}, 34\penalty0 (2):\penalty0
  304--313, 2018.
\newblock ISSN 0267-3649.
\newblock \doi{https://doi.org/10.1016/j.clsr.2017.08.007}.
\newblock URL
  \url{https://www.sciencedirect.com/science/article/pii/S0267364917302091}.

\bibitem[Warnecke et~al.(2021)Warnecke, Pirch, Wressnegger, and
  Rieck]{warnecke2021machine}
Alexander Warnecke, Lukas Pirch, Christian Wressnegger, and Konrad Rieck.
\newblock Machine unlearning of features and labels.
\newblock \emph{arXiv preprint arXiv:2108.11577}, 2021.

\bibitem[Xu et~al.(2023)Xu, Zhu, Zhang, Zhou, and Yu]{xu2023machine}
Heng Xu, Tianqing Zhu, Lefeng Zhang, Wanlei Zhou, and Philip~S Yu.
\newblock Machine unlearning: A survey.
\newblock \emph{ACM Computing Surveys}, 56\penalty0 (1):\penalty0 1--36, 2023.

\bibitem[Zhang et~al.(2023)Zhang, Nakamura, Isohara, and Sakurai]{Zhang2023}
Haibo Zhang, Toru Nakamura, Takamasa Isohara, and Kouichi Sakurai.
\newblock A review on machine unlearning.
\newblock \emph{SN Computer Science}, 4\penalty0 (4):\penalty0 337, Apr 2023.
\newblock ISSN 2661-8907.
\newblock \doi{10.1007/s42979-023-01767-4}.
\newblock URL \url{https://doi.org/10.1007/s42979-023-01767-4}.

\end{thebibliography}
\bibliographystyle{iclr2025_conference}

\appendix
\section{Appendix}

\subsection{Initial Performances of the Models}

We train both: the models with a DKVB and the baseline models (i.e. backbone + linear layer) to achieve similar performances on the test datasets in order to ensure a fair comparison. Note that, due to these the models are not necessarily trained to achieve the maximum possible performance on the datasets. Table \ref{tab:init_perf} shows the initial performances of the originally trained models on different splits of the datasets.

\begin{table*}[h!]
\tiny
\caption{Performance of the models on different sets of data after the initial training on the four datasets. We use two kinds of models: (a) models having a Discrete KV Bottleneck which are used for the proposed methods and (b) models where the DKVB and the decoder are replaced by a Linear Layer. These are used for the baseline. We wish to reduce the accuracy of these models on \({D}_{test}^{forget}\) to 0\% while maintaining the accuracy on \(\mathcal{D}_{test}^{retain}\).}
\label{tab:init_perf}
\centering
\subfigure[\textbf{Backbone + DKVB}]{
\resizebox{\textwidth}{!}{%
\begin{tabular}{c|ccc|ccc|ccc|ccc}
\toprule
 & \multicolumn{6}{c|}{ViT-B/32} & \multicolumn{6}{c}{ResNet-50} \\
\midrule
\textbf{Dataset}  & \textbf{\(\mathcal{D}_{train}\)}  & \textbf{\(\mathcal{D}_{train}^{retain}\)}  & \textbf{\(\mathcal{D}_{train}^{forget}\)}  & \textbf{\(\mathcal{D}_{test}\)}  & \textbf{\(\mathcal{D}_{test}^{retain}\)}  & \textbf{\(\mathcal{D}_{test}^{forget}\)} & \textbf{\(\mathcal{D}_{train}\)}  & \textbf{\(\mathcal{D}_{train}^{retain}\)}  & \textbf{\(\mathcal{D}_{train}^{forget}\)}  & \textbf{\(\mathcal{D}_{test}\)}  & \textbf{\(\mathcal{D}_{test}^{retain}\)}  & \textbf{\(\mathcal{D}_{test}^{forget}\)} \\
\midrule

CIFAR-10                          &100\%                           &100\%                           &100\%                            &93.01\%                          &92.61\%                           &96.50\%    &                         100\%               &           100\%              &           100\%              &          82.94\%                &           82.04\%                &           91.00\%            \\
CIFAR-100                         &99.98\%                           &99.98\%                           &100\%                              &78.43\%                          &78.24\%                           &96.00\%  &              99.98\%             &             99.98\%              &                100\%              &           62.11\%            &           61.81\%                &        92.00\%             \\
LACUNA-100                        &98.09\%                           &98.07\%                           &100\%                              &90.38\%                          &90.28\%                           &100\%           &              98.35\%             &               98.34\%            &          100\%                    &            65.53\%              &               65.24\%            &          94.00\%                \\
ImageNet-1k                        &            99.53\%                   &               99.53\%                 &               100\%                    &                68.24\%               &                68.22\%                & 92.00\% &                99.37\%                 &              99.36\%                    &           100\%                        &                  76.44\%               &              76.41\%                    &     100\%  
                    \\
\bottomrule
\end{tabular}%
}
}

\vspace{1em}

\subfigure[\textbf{Backbone + Linear Layer}]{
\resizebox{\textwidth}{!}{%
\begin{tabular}{c|ccc|ccc|ccc|ccc}
\toprule
 & \multicolumn{6}{c|}{ViT-B/32} & \multicolumn{6}{c}{ResNet-50} \\
\midrule
\textbf{Dataset}  & \textbf{\(\mathcal{D}_{train}\)}  & \textbf{\(\mathcal{D}_{train}^{retain}\)}  & \textbf{\(\mathcal{D}_{train}^{forget}\)}  & \textbf{\(\mathcal{D}_{test}\)}  & \textbf{\(\mathcal{D}_{test}^{retain}\)}  & \textbf{\(\mathcal{D}_{test}^{forget}\)} & \textbf{\(\mathcal{D}_{train}\)}  & \textbf{\(\mathcal{D}_{train}^{retain}\)}  & \textbf{\(\mathcal{D}_{train}^{forget}\)}  & \textbf{\(\mathcal{D}_{test}\)}  & \textbf{\(\mathcal{D}_{test}^{retain}\)}  & \textbf{\(\mathcal{D}_{test}^{forget}\)} \\
\midrule
CIFAR-10                          &93.27\%                           &92.82\%                           &97.32\%                            &93.02\%                          &92.59\%                           &96.90\% &            83.80\%              &           83.42\%              &           87.20\%              &          82.04\%                &           81.74\%                &           84.70\%           \\
CIFAR-100                         &86.73\%                           &86.61\%                           &99.00\%                              &78.53\%                          &78.35\%                           &96.00\% &               78.02\%            &           77.82\%               &                  98.20\%            &           62.69\%               &            62.41\%             &            90.00\%         \\
LACUNA-100                        &95.58\%                           &95.53\%                           &100\%                              &90.68\%                          &90.59\%                           &100\%     &               86.48\%            &              84.53\%             &              99.25\%                &              65.40\%            &               65.10\%            &                95.00\%          \\
ImageNet-1k                       &            73.13\%                      &              73.11\%                    &                 96.15\%                  &             68.29\%                    &        68.26\%
              & 92.00\% &              97.44\%                      &              97.43\%                  &             100\%                       &          76.77\%                      &        76.75\%
              &              100\%                 \\
\bottomrule
\end{tabular}%
}
}
\end{table*}

\subsection{Deciding the Forget Class}
\label{app:forget}
\begin{figure}[h]
     \centering
     \subfigure[]{
         \includegraphics[width=4.4cm, height=4.4cm]{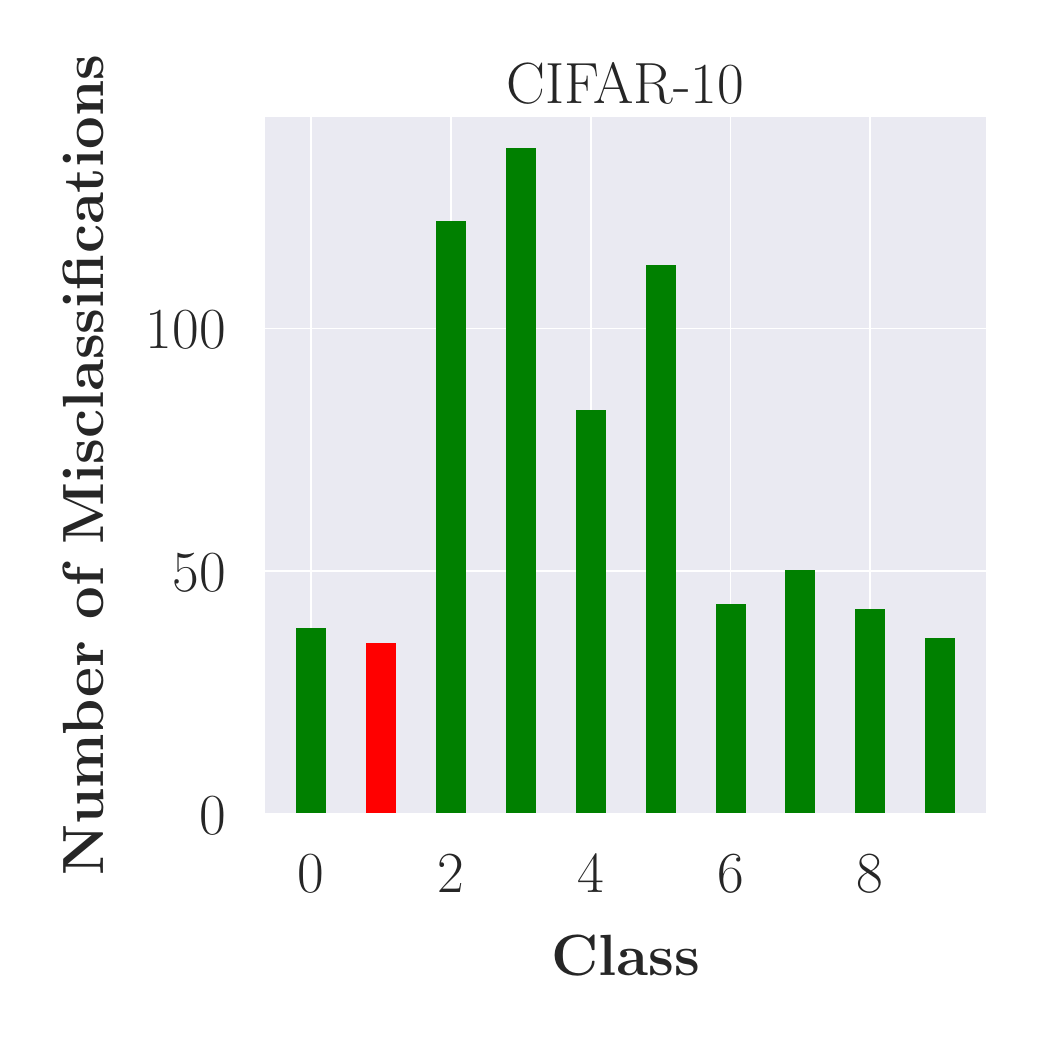}
         \label{fig4:a}    
     }
     \subfigure[]{
         \includegraphics[width=4.4cm, height=4.4cm]
         {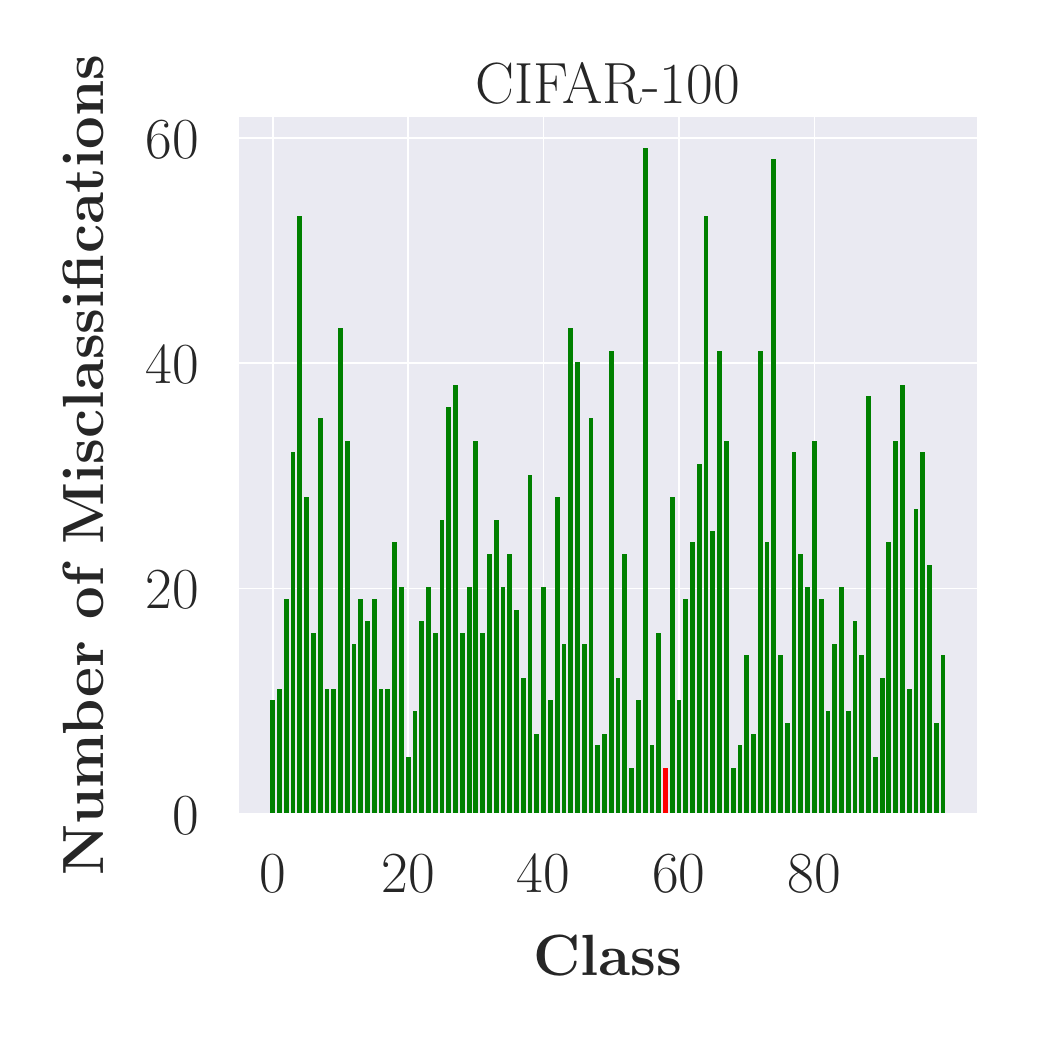}
         \label{fig4:b}
     }
     \subfigure[]{
        \includegraphics[width=4.4cm, height=4.4cm]
        {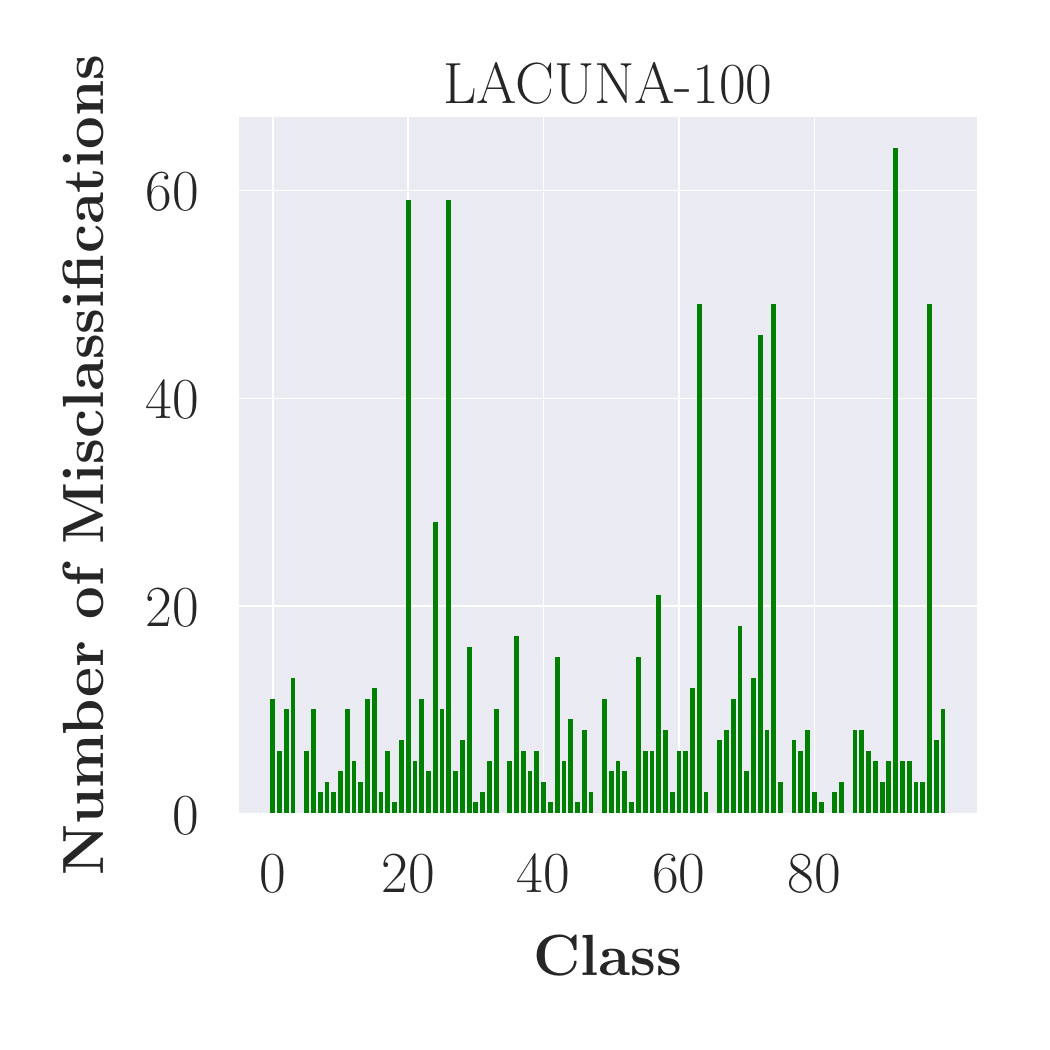}
        \label{fig4:c}
     }
        \caption{Number of mis-classifications per class for the test data. The red bars correspond to the class with the least number of mis-classifications (a) CIFAR-10: Class 1 has the least number of mis-classifications (b) CIFAR-100: Class 58 has the least number of mis-classifications (c) LACUNA-100: Classes 34, 48, 65, 76, 82 and 85 have 0 mis-classifications and hence, do not have a bar}
\end{figure}

We assume that this class should be the most difficult one for the model to forget. Figures \ref{fig4:a} - \ref{fig4:c} show the number of mis-classifications per class on the test data, for CIFAR-10, CIFAR-100 and LACUNA-100 for the ViT-B/32 backbone. For CIFAR-10, class \#1 is the best-learned class with the lowest number of mis-classifications. Thus, we select class \#1 as the forget class for the dataset. For CIFAR-100 class 58 is the best-learned class and for LACUNA-100, class 48 is one of the best-learned classes with zero mis-classifications. Hence, we select classes \#58 and \#48 as the forget classes for CIFAR-100 and LACUNA 100 respectively. We determine the forget class in all other cases using the same method. We use the same forget classes for experiments on the models with a linear layer in place of the DKVB (i.e., the baseline) as well. Table \ref{tab:forget_classes} shows the forget class for all the cases discussed in our experiments.

\begin{table*}[h!]
\caption{Forget classes for the different scenarios presented in the paper}
\begin{center}
\begin{tabular}{c|cc|cc|cc|cc}
\toprule
Forget Classes & \multicolumn{2}{c}{CIFAR-10}   & \multicolumn{2}{c}{CIFAR-100} & \multicolumn{2}{c}{LACUNA-100} & \multicolumn{2}{c}{ImageNet-1k} \\
\midrule
ViT-B/32  &       \multicolumn{2}{c|}{1}       &     \multicolumn{2}{c|}{58}       &         \multicolumn{2}{c|}{48}     &         \multicolumn{2}{c}{1}           \\
ResNet-50     &    \multicolumn{2}{c|}{1}      &   \multicolumn{2}{c|}{94}   &  \multicolumn{2}{c|}{34}   &  \multicolumn{2}{c}{9}               \\
\bottomrule
\end{tabular} 
\label{tab:forget_classes}
\end{center}
\end{table*}

\subsection{Multi Class Unlearning}
\label{app:multi_unlearning}
\begin{figure}[h]
    \centering
    \subfigure[]{
        \includegraphics[width=4.4cm, height=4.4cm]{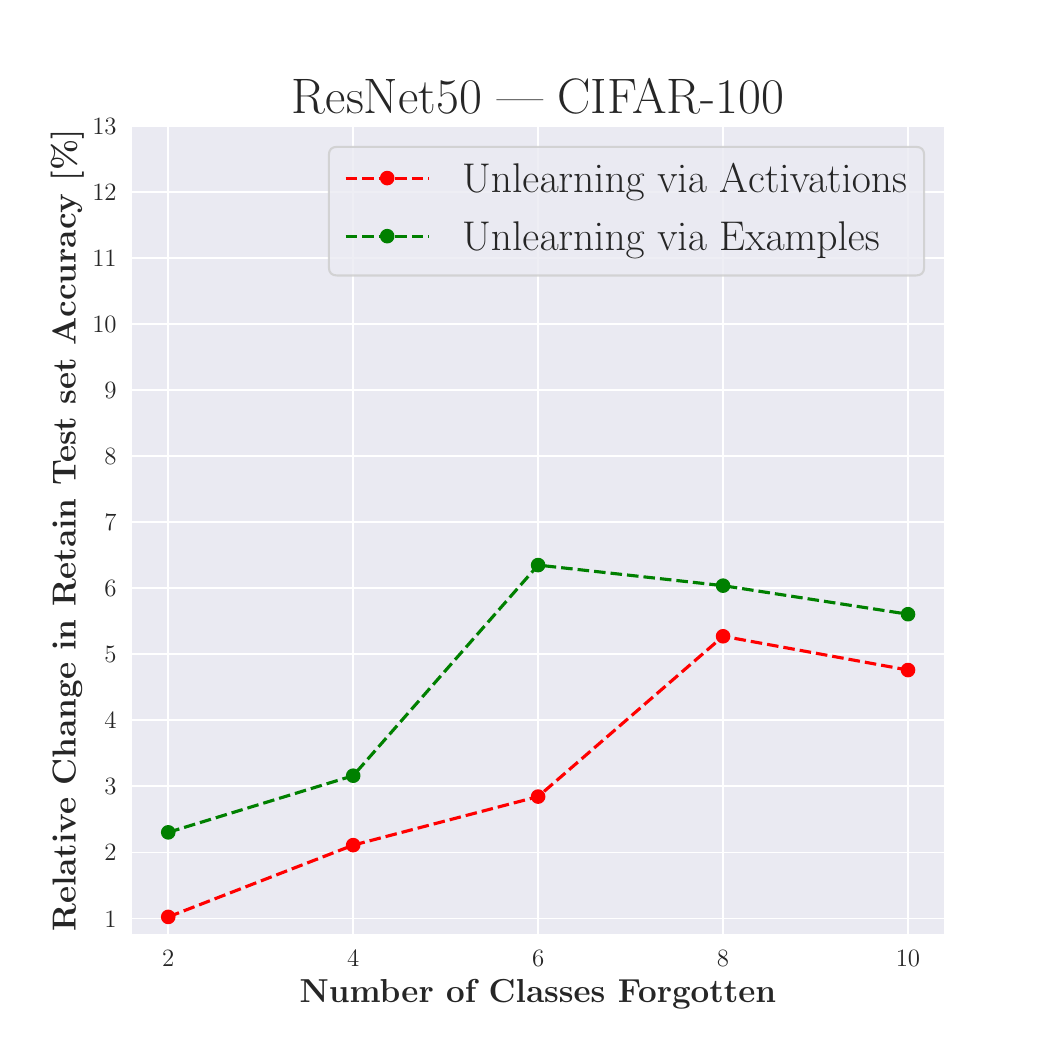}
    }
    \subfigure[]{
        \includegraphics[width=4.4cm, height=4.4cm]
        {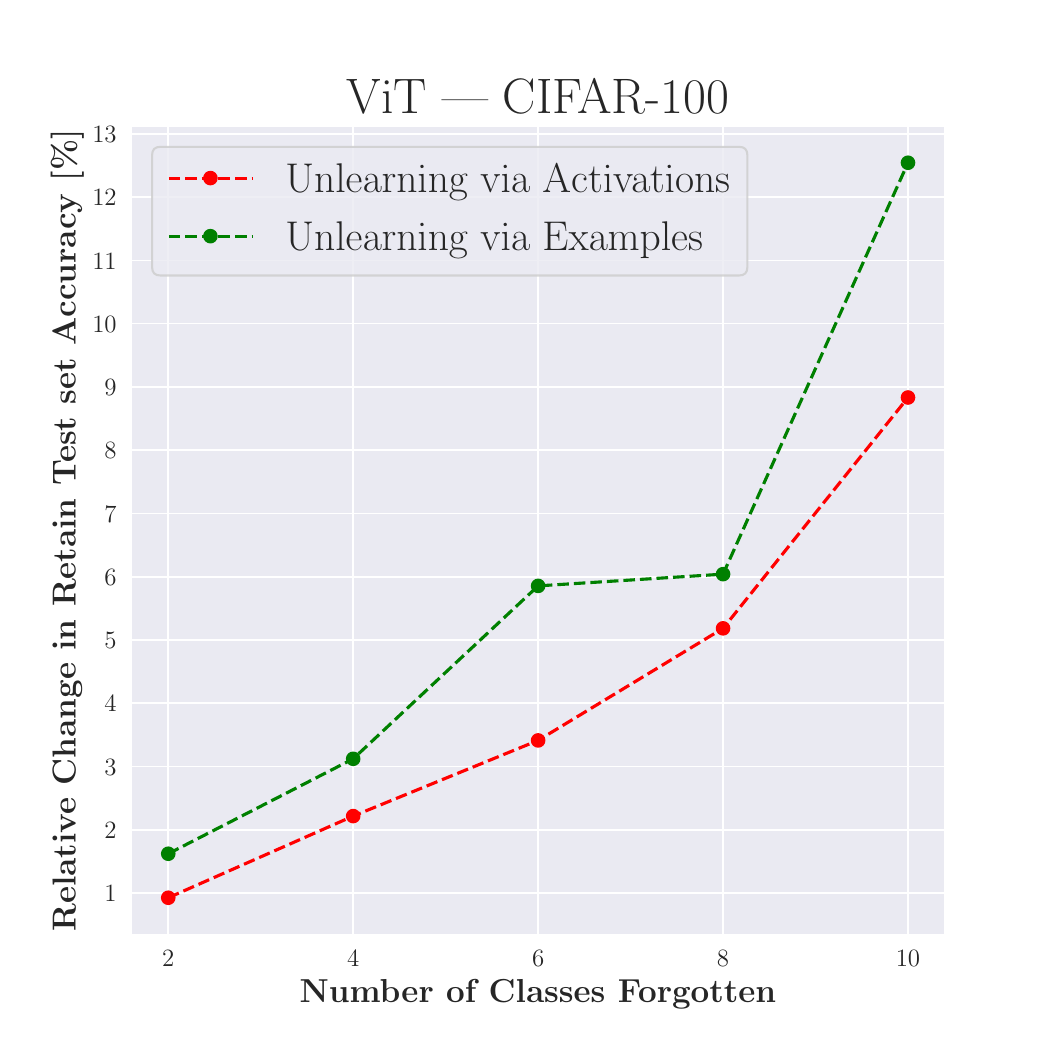}
    }
    \label{fig:multi_unlearning}
    \caption{Multi Class Unlearning for CIFAR-10}
\end{figure}

We attempt to investigate the effects of unlearning multiple classes at once by performing experiments on CIFAR-100 for both ViT/B-32 and ResNet-50 models. We unlearn upto 10 classes using both - Unlearning via Activations as well as Unlearning via Examples, and run each experiment for 5 seeds. The classes to be forgotten are chosen randomly. The given below show the unlearnt model's performance on the retain class vs the number of classes unlearnt at approximately the point of complete unlearning. Figure~\ref{fig:multi_unlearning} shows the results of this experiment.

We can see that a ResNet-50 backbone performs comparatively better. Also, Unlearning via Activations performs relatively better as compared to Unlearning via Examples. We can also clearly see that the relative error as compared to the original model on the retain set performance are higher than single class unlearning, as expected. Noticeably, the performance starts to significantly deteriorate when forgetting > 6 classes in the case of Unlearning via Examples, and > 8 examples in the case of Unlearning via Activations for both the backbones. 

\subsection{Choosing the forget class randomly}
\label{app:forget_random}
To ensure that the effectiveness of the approach is not class specific, we perform experiments for CIFAR-100, where the class to be forgotten is randomly chosen, and compare the performance of the proposed approaches against SCRUB [2]. We run each experiment for 5 random seeds, wherein the forget class is randomly chosen for each seed. Rest of the experimental setup remains the same as described in Section 5 of the paper. We report the results in Table~\ref{tab:forget_random}

\begin{table*}[t]
    \caption{Selecting the forget class randomly for CIFAR-100}
    \label{tab:forget_random}
    \begin{center}
    \resizebox{\textwidth}{!}{
    \begin{tabular}{c|c|cc}
    \toprule
                                        & & \multicolumn{2}{c}{CIFAR-100} \\
    \midrule
                                Backbone & Method & $\mathcal{D}_{test}^{retain}$   & $\mathcal{D}_{test}^{forget}$ \\
    \midrule
    \multirow{3}{*}{ViT/B-32} & DKVB via Activations (sec \ref{subec:results_zeroshot})  &  -0.27 $\pm$ 0.07 \%      & -100\% $\pm$ 0 \%              \\
    &DKVB via Examples (sec \ref{subec:results_zeroshot})     &  -0.47 $\pm$ 0.03 \%      &-100 $\pm$ 0 \%      \\
    \midrule
    \multirow{3}{*}{ResNet-50} & DKVB via Activations (sec \ref{subec:results_zeroshot})  &    0.58 $\pm$ 0.12 \%   &   -100 $\pm$ 0 \%           \\
    &DKVB via Examples (sec \ref{subec:results_zeroshot})     &    0.28 $\pm$ 0.11 \%   &   -100 $\pm$ 0 \%     \\
    \bottomrule
    \end{tabular}
    }              
    \end{center}
    \vspace{-0.3cm}
    \end{table*}

Clearly, even with the forget class chosen randomly, the proposed approaches perform equally well as in the scenario where the forget class is fixed. This proves that the effectiveness of the proposed approaches is not class dependent.

\subsection{Unlearning beyond the compute free setting}
\label{app:results_relearning}

\begin{figure}[h]
     \centering
     \subfigure[CIFAR-10]{
         \begin{minipage}{0.3\textwidth}
            \includegraphics[width=\linewidth]{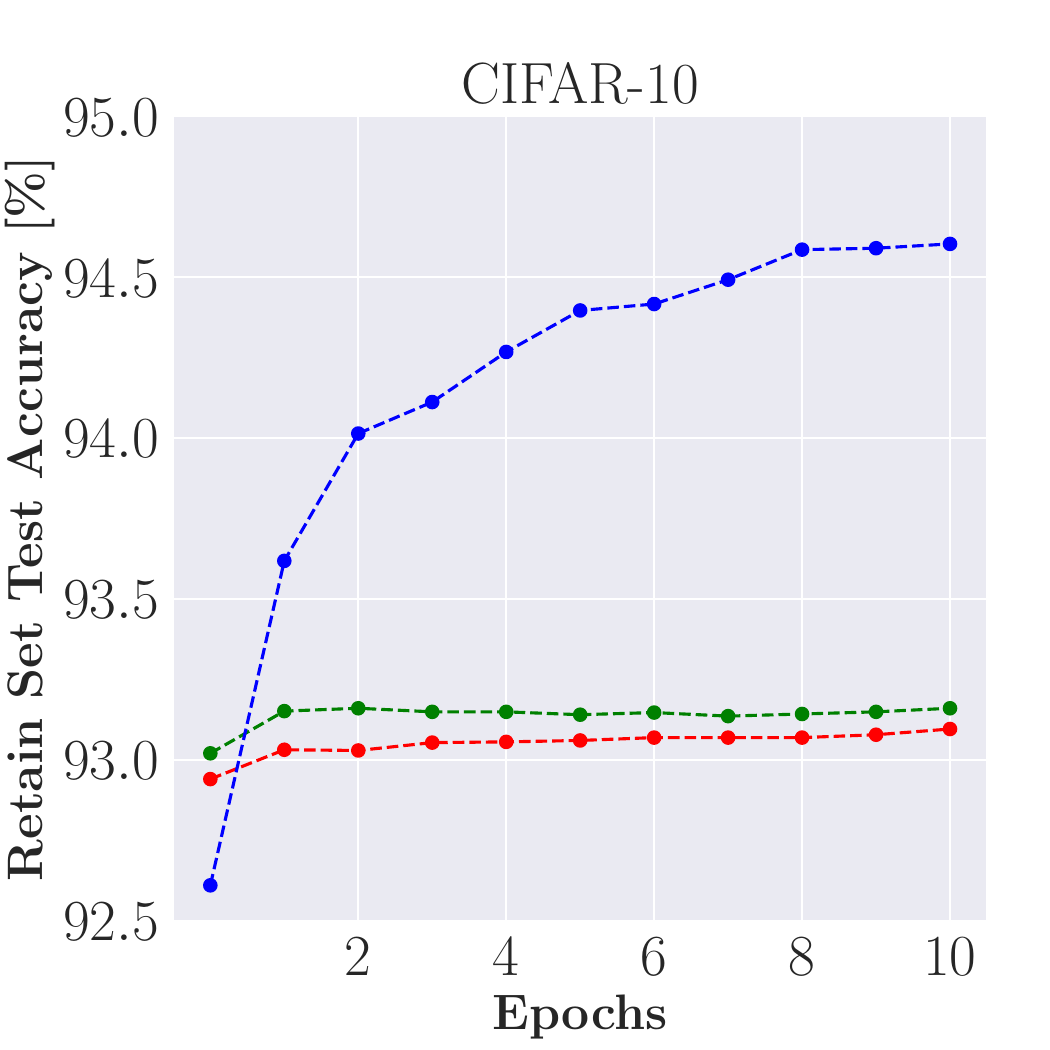}
        \end{minipage}
         \label{fig5:a}    
     }
     \subfigure[CIFAR-100]{
         \begin{minipage}{0.3\textwidth}
            \includegraphics[width=\linewidth]{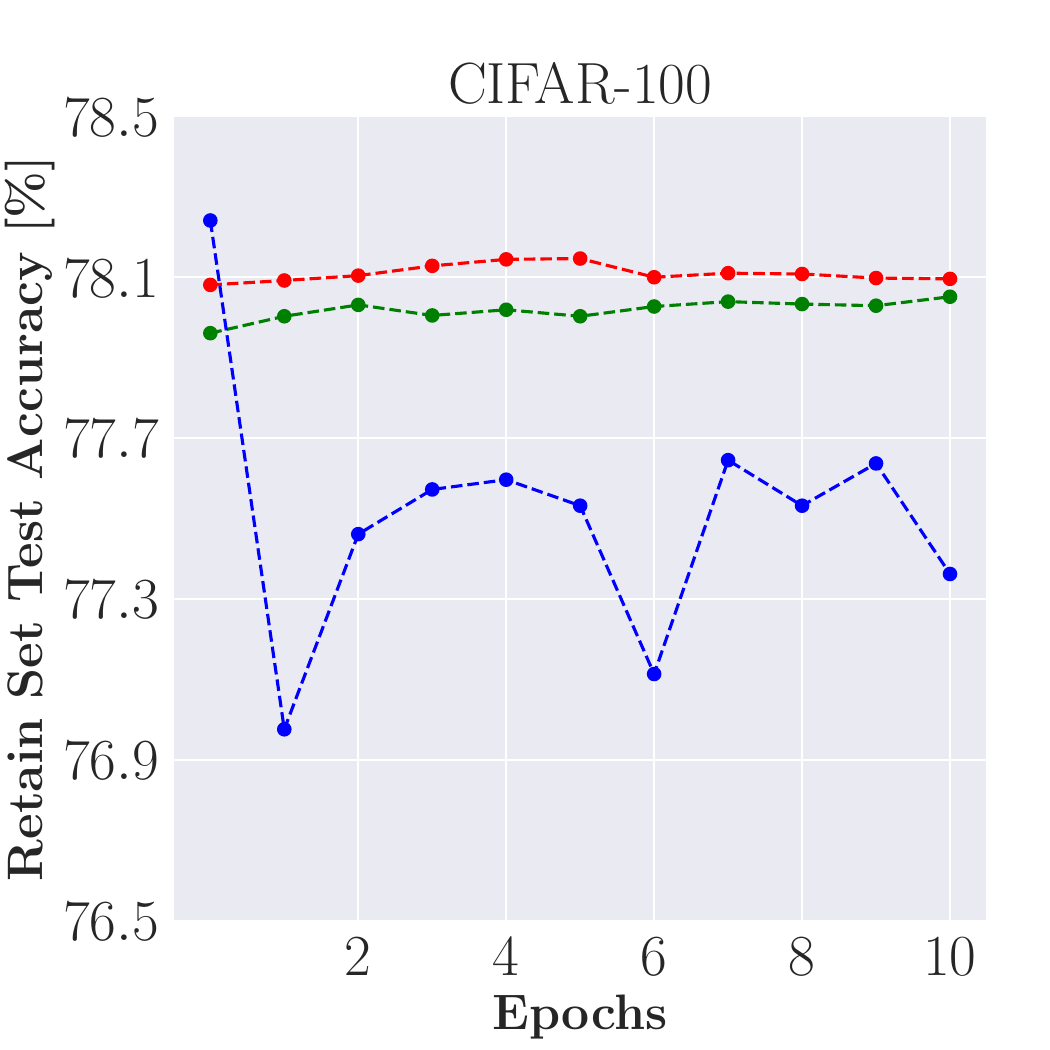}
        \end{minipage}
         \label{fig5:b}
     }
     \subfigure[LACUNA-100]{
        \begin{minipage}{0.3\textwidth}
            \includegraphics[width=\linewidth]{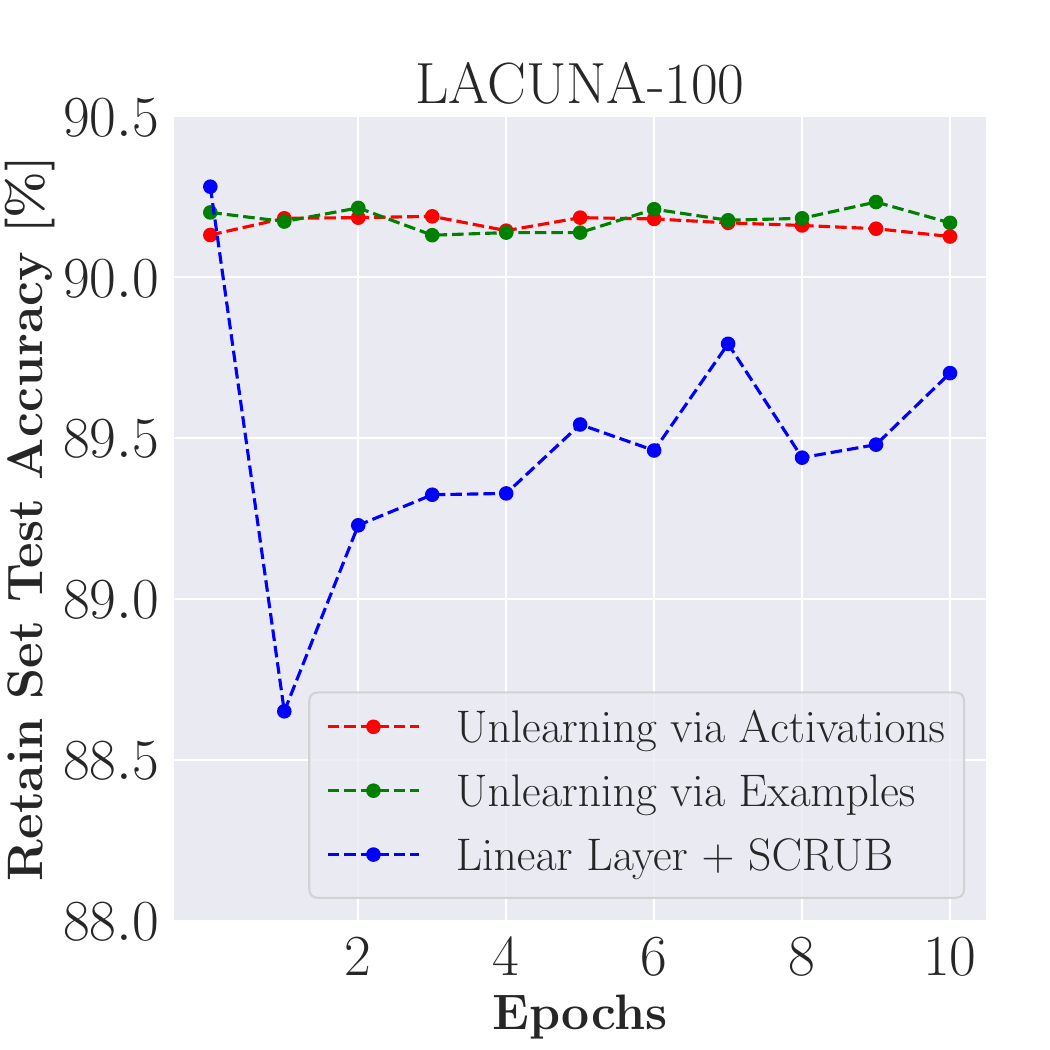}
        \end{minipage}
        \label{fig5:c}
     }
        \caption{Comparison between the performance of proposed methods with added compute and the baseline on the retain set test data. For the proposed methods, the plots start from after the initial zero shot unlearning. For the baseline, the plots start from the original models. Retraining the models unlearned using the proposed models does not lead to any significant improvements in performance.}
        \label{fig:5}
\end{figure}

\begin{figure}[h]
     \centering
     \subfigure[CIFAR-10]{
         \begin{minipage}{0.3\textwidth}
            \includegraphics[width=\linewidth]{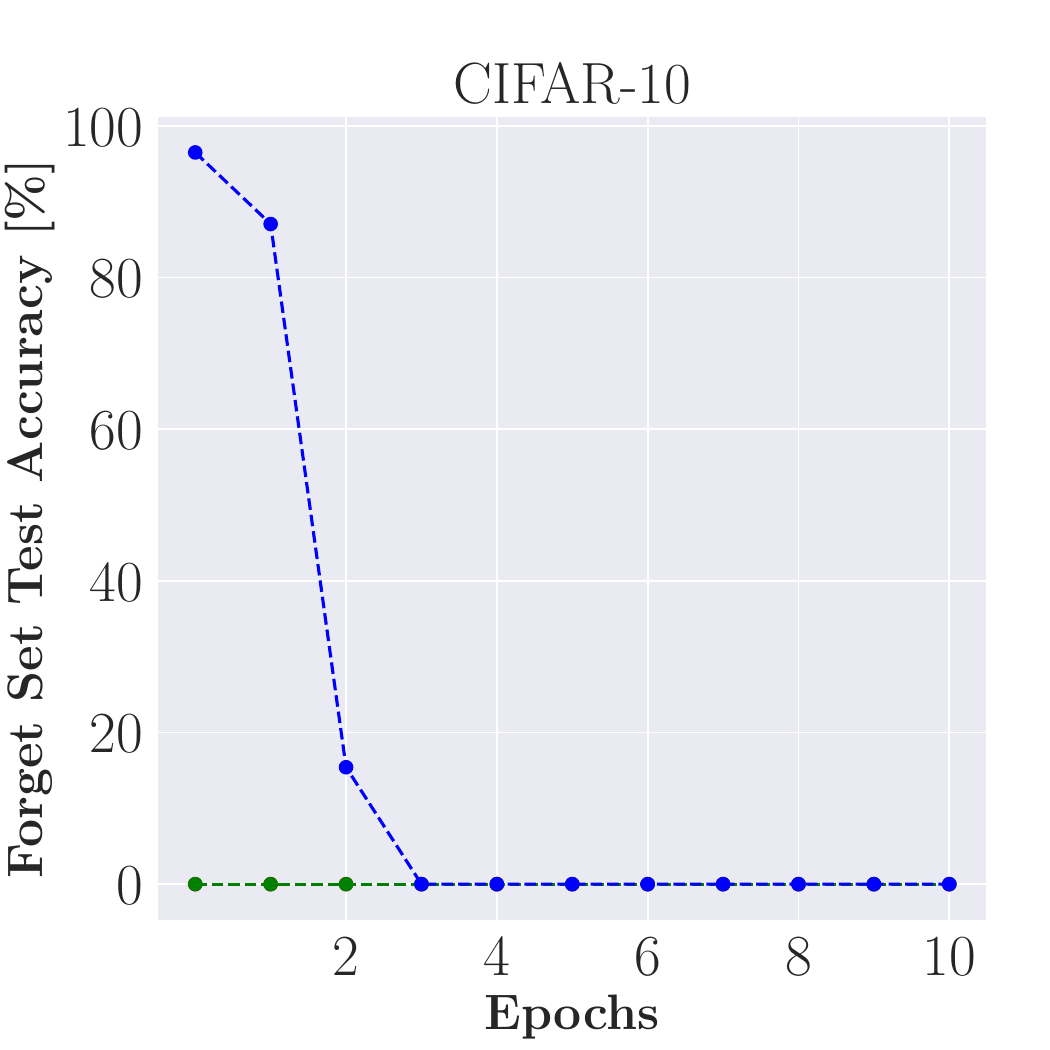}
        \end{minipage}
         \label{fig6:a}    
     }
     \subfigure[CIFAR-100]{
         \begin{minipage}{0.3\textwidth}
            \includegraphics[width=\linewidth]{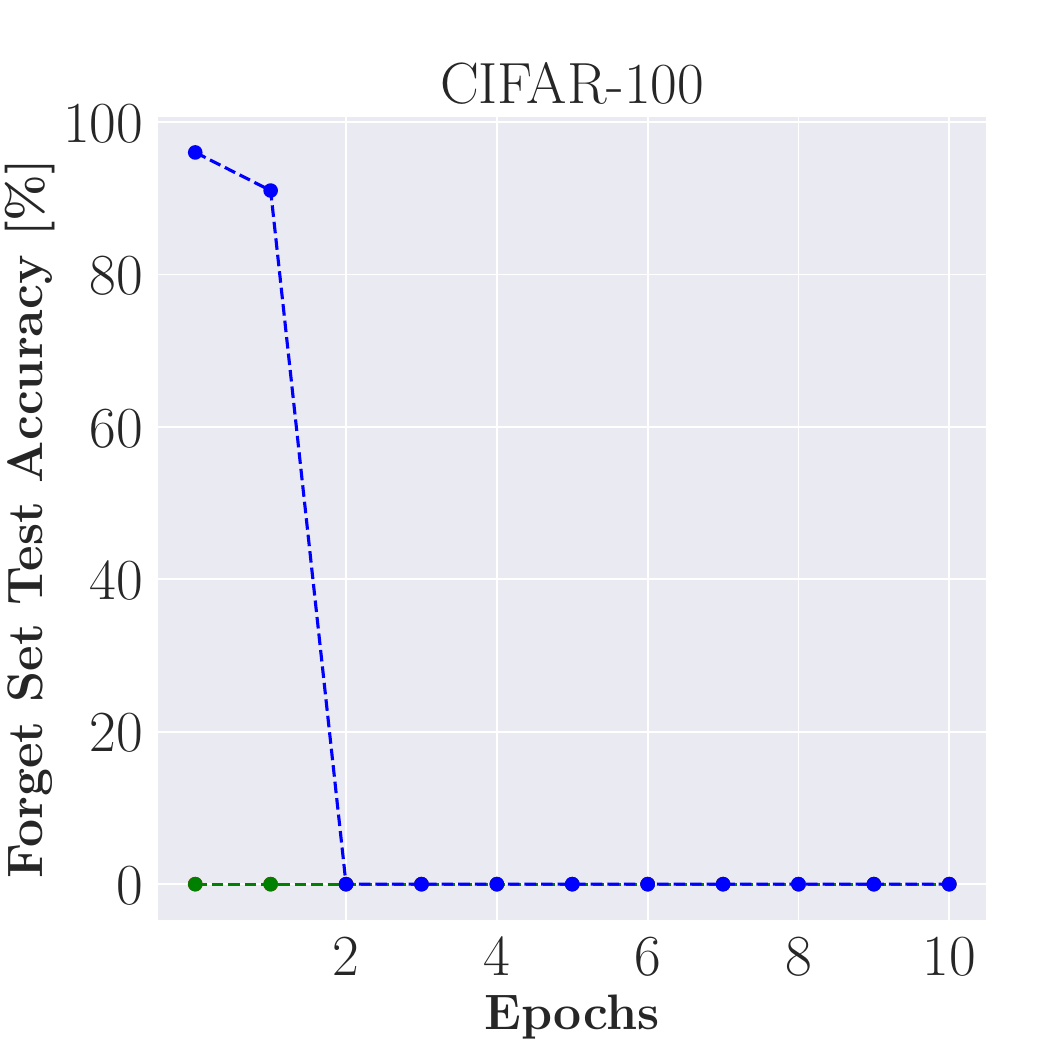}
        \end{minipage}
         \label{fig6:b}
     }
     \subfigure[LACUNA-100]{
        \begin{minipage}{0.3\textwidth}
            \includegraphics[width=\linewidth]{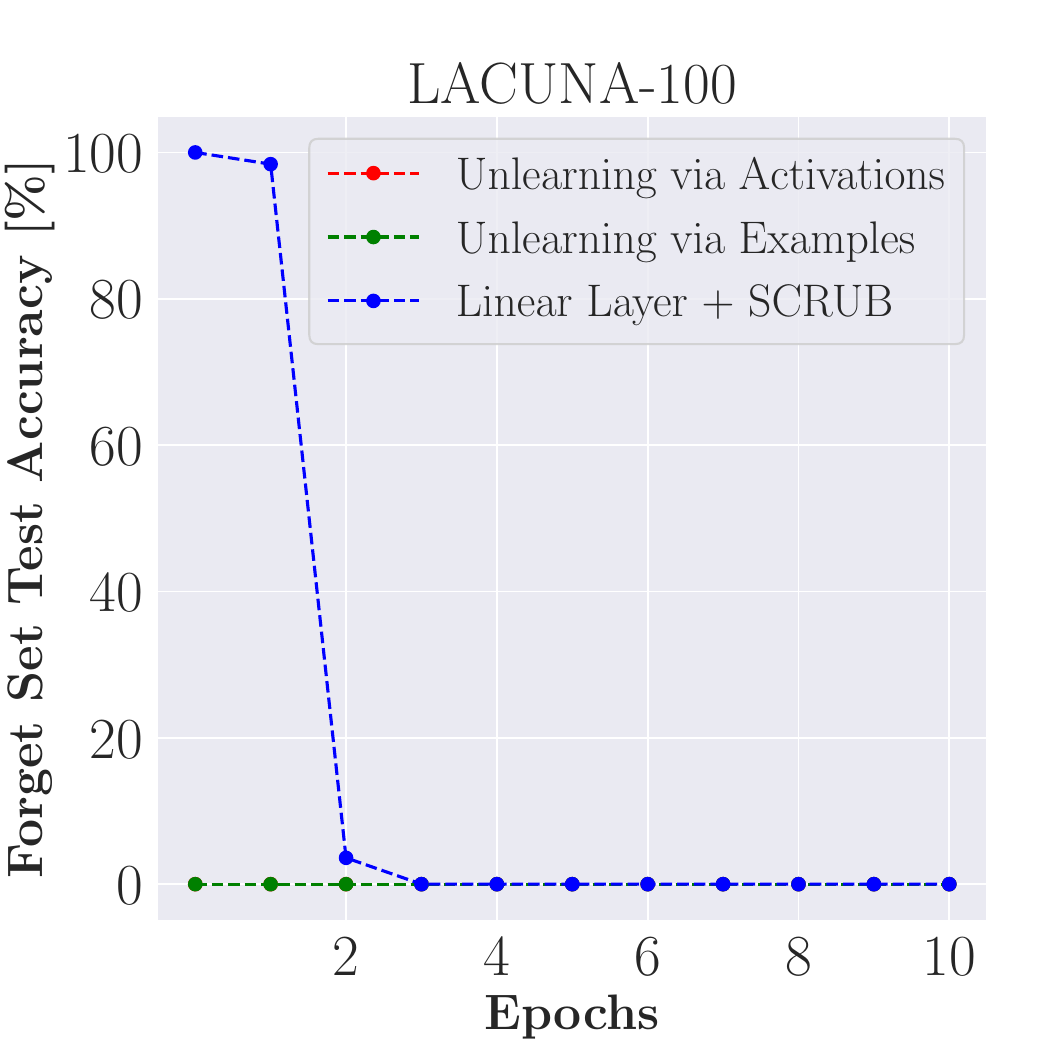}
        \end{minipage}
        \label{fig6:c}
     }
        \caption{Comparison between the performance of proposed methods with added compute and the baseline on the and forget set test data. Note that for the proposed methods, the plots start from after the initial zero shot unlearning. For the baseline, the plots start from the original models. The green line occludes the red line since both of them stay at 0\% throughout the training.}
        \label{fig:6}
\end{figure}

We investigate the effect of using additional compute to the proposed methods. As shown previously, the proposed methods perform competitively to SCRUB. To the best of our knowledge, SCRUB is the most competitive and relevant unlearning approach. 
However, it has the inherent drawback of requiring compute for unlearning. Nevertheless, for a fair comparison, we additionally explore the implications of this additional compute for the proposed two methods for a ViT/B-32 backbone on CIFAR-10, CIFAR-100 and LACUNA-100. 
Specifically, we retrain the DKVB models after the (compute efficient) unlearning, on the training data of the retain set (i.e., $\mathcal{D}_{train}^{retain}$) for 10 epochs. 
For the baseline, we use the same experimental setting as in Section~\ref{sec:compbase}, except - we run it for 10 epochs instead of stopping when either the forget set has been completely unlearned or the performance has converged.
Figure~\ref{fig:5} and figure~\ref{fig:6} highlight the effect of retraining of the proposed methods compared to SCRUB across multiple epochs, for all three datasets. 

Retraining the unlearned models on the retain set does not affect their performance significantly.
The performance of the baseline on the other hand increases after an initial drop in case of CIFAR-100 and LACUNA-100. The initial drop may be attributed to the damage to the retain set performance caused by the initial \textit{max-steps}. The subsequent increase can be attributed to the fact that the SCRUB training objective also optimizes the task loss on the retain set. Thus, once the model unlearns the forget set, SCRUB shifts the model capacity towards better learning the retain set. For CIFAR-10 this results in the model performing better than the DKVB models on the retain set as the retain set test accuracy after unlearning is higher than the original model. However, the baseline is unable to recover its original performance for CIFAR-100 and LACUNA-100.

For the forget set, in all three cases, the baseline completely unlearns the forget set quickly within the first few epochs, as shown in Figure~\ref{fig:6}. 

\label{app:retr_forget}

\subsection{Using the proposed Methods against Membership Inference Attacks}
\label{app:mia}

Depending on the application, complete unlearning of the forget set may not always be the final goal of unlearning. For several use cases such as removing information about corrupted data from the model or removing  harmful biases exhibited by the model, maximal error on the forget set is desirable. However, for applications such as Differential Privacy, it is more desirable to achieve a forget set error which is similar to that of a model trained from scratch only on the retain set. Otherwise, it makes the unlearned model susceptible to Membership Inference Attacks (MIA) \citep{shokri2017membership}. Although we do not explore this setting in detail in this work, the proposed method can also be used for applications where complete unlearning is not desirable. This can be done by following a procedure similar to SCRUB+R \citep{kurmanji2023towards}, wherein instead of selecting a particular model checkpoint, one can select the model corresponding to particular values of $N_a$ or $N_e$ such that the error on the forget set test data is similar to the reference point as defined in \citet{kurmanji2023towards}. 

First,it is important to clarify that the proposed approaches are not a-priori suited for selective unlearning, i.e. the setting where we want the model to forget specific examples or a small subset of examples instead of removing the information about an entire class. The KV bottleneck essentially induces induces clusters of representation, where the members of a particular cluster correspond to the representations belonging to the same class (see Figure 2 in \citet{trauble2023discrete}). When we try to unlearn the representations corresponding to one particular example belonging to a particular class, the KV bottleneck routes the selection to other (key-)representations within the same cluster since those keys would be the next closest to the representation of the encoder. Since these representations also contain information about the same class as the examples we intend to unlearn, the model would still predict the class to be unlearnt.

Due to the same reason the proposed approaches are also not designed for working against traditional Membership Inference attacks. According to the basic attacks setup as explained in \citet{kurmanji2023towards}, the objective is to obtain a model that has unlearnt a small subset of specific examples (i.e. selective unlearning) such that the loss of the model on the unlearnt subset of examples should be indistinguishable from loss on examples that the model never saw during training.

Nevertheless, since the proposed approaches are designed for class unlearning specifically, we attempt to evaluate them on a modified version of the above. We call this "Class Membership Inference Attacks (CMIA)". In CMIA, the aim is to defend against an attacker whose aim is to determine whether a model that has undergone unlearning ever saw a particular class as a part of its training data. Thus, we want the model to unlearn a particular class such that the losses/performance of the model on the unlearnt class as a whole, is indistinguishable from those on a held-out class that the model never saw during its training. We describe the experimental setup and results below.

\textbf{Experimental Setup} We perform the experiment for CIFAR10 with a ViT/B-32 backbone. We divide the dataset into training data ($D_{Train}$), validation data ($D_{Val}$) and test data ($D_{Test}$). Training Data consists of 4000 examples per class; validation and test data consist of 1000 examples per class. We first trained a model on the first 9 classes of CIFAR10. Thus, class number 10 is the held-out class. Next, we unlearn class 1 from the model using the Unlearning via Activations approach introduced in the paper. We unlearn the model until the loss of the model on the validation sets of the forget class and the held-out class are similar. In our experiments, we find that we reach this point at approximately $N_a = 240000$. The loss $l(x, y)$ in our case is be the cross-entropy loss.

Next, we label the losses corresponding to the validation and test set of the forget class as 1 and those corresponding to the validation and test set of the held-out class as 0. We train a binary classifier on the validation losses of the forget and held-out sets and evaluate it on the test losses. We follow a similar setting for the baseline model, where we obtain the model suitable for MIA defense by using SCRUB+R \citep{kurmanji2023towards}. For a successful defense, we would want the accuracy of the classifier to be close to 50\% on the test losses, indicating that it is unable to distinguish between the unlearned class and the held-out class. Same as \citet{kurmanji2023towards}, we use \texttt{sklearn.logistic\_regression} as our attacker (the binary classifier). We call the approach described above Partial UvA (Partial Unlearning via Activations). We run experiments for 3 random seeds, and the mean of the attacker performance is reported. Note that a similar procedure can also be followed using Unlearning via Examples.

\textbf{Observations and Results}: We report the results of the experiment described above in the table given below. We observe that although the baseline performs slightly better, the proposed approaches perform competitively, even though we have not intended to develop the method for this scenario.

\begin{table}[h]
\caption{Comparison on Class Membership Inference Attacks between the proposed approach and the baseline. A binary classifier is trained on the validation losses of the forget and held-out sets and is evaluated on the test losses. The proposed approach performs competitively to SCRUB + R}
\label{tab:mia}
\begin{center}
\begin{tabular}{c|c}
\toprule
                           Approach & Attacker Accuracy   \\
\midrule
Partial UvA  &  53.50\%            \\
Linear Layer + SCRUB + R     &  \textbf{51.50\%}      \\
\bottomrule
\end{tabular}              
\end{center}
\end{table}

\subsection{Training Details and Hyperparameters}

We perform all of our experiments on a 48GB RTX8000 GPU. We do not use any data augmentation in any experiment. The transforms used for training the model with a ViT/B-32 backbone are the same as CLIP \citep{radford2021learning} pretrained ViT/B-32 transforms. For ResNet-50, both pre-trained weights and transforms are loaded from \texttt{torchvision.models.ResNet50\_Weights}

\subsubsection{Training Details and Hyperparameters for training the original DKVB Models}
\label{hp:dkvb}
In the case of ImageNet pretrained ResNet-50, the representations of the backbone are extract from the 3rd last layer for CIFAR-10, CIFAR-100 and LACUNA-100 and from the 4th last layer for ImageNet-1k. Table~\ref{tab:hp_base_dkvb} shows all the hyperparameters used for training the base DKVB models.
\vspace{-0.4cm}
\begin{table}[h]
\tiny
\caption{Hyperparameters used for training the base DKVB models}
\label{tab:hp_base_dkvb}
\begin{center}
\begin{tabular}{c|c|cccc}
\toprule
  Backbone & Hyperparameter                          & CIFAR-10   & CIFAR-100  & LACUNA-100 & ImageNet-1k \\
\midrule
\multirow{10}{*}{ViT/B-32} & top-k                       &   1        &   10       &    10   &   1   \\
 & Key Dimension               &   8        &   8        &    8      &  14  \\
 & \# of Key Init Epochs       &   10       &   10       &    10   &  10   \\
 & Type of Value Init          &   Gaussian Random   &      Zeros    &  Uniform Random  &  Zeros \\
 & \# of Codebooks             &       256             &     256        &   256  &  256  \\
 & \# of Key-Value Pairs per Codebook   &    4096     &  4096       &   4096 &  4096  \\
 & Optimizer             &     Adam       &    Adam    &   Adam & Adam \\
 & LR                   &      0.1       &     0.3       &   0.3 & 0.3 \\
 & Batch Size           &     256         & 256          & 256 & 256 \\
 & Epochs               &      74           & 71            &  7 & 3\\
\midrule
\multirow{10}{*}{ResNet-50} & top-k                       &     1      &    2      &    1  &  1     \\
 & Key Dimension               &      14     &     14     &    8  &   14    \\
 & \# of Key Init Epochs       &     10     &       10   &    10   &   10  \\
 & Type of Value Init          &   Zeros   &      Random    & Gaussian Random   & Gaussian Random   \\
 & \# of Codebooks             &         256          &      256       &   256   & 256 \\
 & \# of Key-Value Pairs per Codebook   &  4096       &    4096    &   4096  &  4096\\
 & Optimizer             &      Adam      &    Adam    &  Adam  & Adam \\
 & LR                   &       0.3      &         0.3  &  0.1 & 0.3 \\
 & Batch Size           &       256       &      256     & 256 & 256 \\
 & Epochs               &       70          &      4       &  1  & 5 \\
\bottomrule
\end{tabular}              
\end{center}
\end{table}

\subsubsection{Training Details and Hyperparameters for training the original Baseline Models}
\label{hp:base}
For the baseline models, we deliberately train them to similar test ($\mathcal{D}_{test}$) accuracies as the models with a Discrete Key Value Bottleneck to ensure a fair comparison for unlearning. Table~\ref{tab:hp_base_linear} shows the hyperparameters used for training the baseline models.

\begin{table}[h]
\caption{Hyperparameters used for training the baseline models}
\label{tab:hp_base_linear}
\begin{center}
\begin{tabular}{c|c|cccc}
\toprule
Backbone & Hyperparameter & CIFAR-10   & CIFAR-100  & LACUNA-100 & ImageNet-1k\\
\midrule
\multirow{3}{*}{ViT/B-32} & LR                   &      0.001       &     0.01       &   0.01 &  0.01 \\
 & Batch Size           &     256         & 256          & 256 &  512 \\
 & Epochs               &      1           & 7            &  13  &  1 \\
 \midrule
\multirow{3}{*}{ResNet-50} & LR                   &       0.01      &     0.001      &  0.01  & 0.001  \\
 & Batch Size           &       256       &    256       & 512 & 512 \\
 & Epochs               &         2        &    72         &  73  & 11 \\
\bottomrule
\end{tabular}              
\end{center}
\end{table}

\subsubsection{Training Details and Hyperparameters for SCRUB}
\label{hp:scrub}
For the baseline, we run SCRUB on the model with linear layer. One \textit{epoch} consists of one \textit{min step} and may or may not contain a \textit{max step}. Hence the values of \textit{min steps} and \textit{epochs} are always same. One \textit{max step} is included in every epoch for the first \textit{msteps} epochs. We tune the hyperparameter \textit{msteps} in our experiments and pick the case where the model is able to best recover its performance on the retain set test data and consider this model as the final unlearned model. We mention the hyperparameters used for running SCRUB corresponding to the results presented in Section~\ref{sec:compbase} in Table~\ref{tab:hp_scrub}. In this case, training of SCRUB is stopped when the forget set accuracy has either dropped to 0\% or converged at a close to 0\% value without damaging the retain set accuracy. Results presented in Appendix~\ref{app:results_relearning} also use the same set of hyperparameters except \textit{min-step} which is always 10 since we train all the methods for 10 epochs.

\begin{table}[h]
\caption{Hyperparameters for SCRUB + Linear Layer Experiments shown in Section~\ref{sec:compbase}}
\label{tab:hp_scrub}
\small
\begin{center}
\begin{tabular}{c|c|cccc}
\toprule
Backbone &  Hyperparameter  & CIFAR-10   & CIFAR-100  & LACUNA-100 & ImageNet-1k \\
\midrule
\multirow{6}{*}{ViT/B-32} & Forget Set Batch Size    & 256 & 256 & 256 & 512 \\
 & Retain Set Batch Size    & 256  & 256 & 256 & 512 \\
 & \# of max-steps (\textit{msteps})    & 3    & 9 & 5 &   3  \\
 & \# of min-steps / \# of epochs     &    3    &  10   & 7 &  3 \\
 & LR                   &      0.001       &     0.01       &   0.01   & 0.001 \\
 & Optimizer           &        Adam      & Adam & Adam  &  Adam \\
 \midrule
 \multirow{6}{*}{ResNet-50} & Forget Set Batch Size    & 256 & 256 & 256 & 512  \\
  & Retain Set Batch Size    & 256  & 256 & 256 & 512 \\
   & \# of max-steps (\textit{msteps})    & 9    & 3 & 3 &  3   \\
 & \# of min-steps / \# of epochs     &    10    &  30  & 30 &  10 \\
  & LR                   &     0.01        &      0.001      &   0.01   & 0.001 \\
   & Optimizer           &      Adam        & Adam &  Adam & Adam  \\
   
\bottomrule
\end{tabular}              
\end{center}
\end{table}

\subsubsection{Training Details and Hyperparameters for Retraining Experiments}
\label{hp:retraining}
Once the DKVB models are unlearned using \textit{Unlearning via Activations} and \textit{Unlearning via Examples}, we retraining them in order to make a fair comparison with the baseline. Thus, during retraining, the initial performance of these models on the retain set is same as the final performance of the unlearned models. Table~\ref{tab:hp_retrain} show the hyperparameters used for retraining the unlearned DKVB models.

\begin{table}[h]
\caption{Hyperparameters used for re-training experiments. \textbf{UvA} stands for \textit{Unlearning via Activations} and \textbf{UvE} stands for \textit{Unlearning via Examples}}
\label{tab:hp_retrain}
\begin{center}
\begin{tabular}{c|cc|cc|cc}
\toprule
                            & \multicolumn{2}{c}{CIFAR-10}   & \multicolumn{2}{c}{CIFAR-100} & \multicolumn{2}{c}{LACUNA-100} \\
\midrule
                     &  UvA & UvE & UvA & UvE & UvA & UvE \\
\midrule
LR                   &   0.3   & 0.3  & 0.1  & 0.1 & 0.1 & 0.3 \\
Optimizer            &    Adam  & Adam & Adam & Adam & Adam & Adam \\
Batch Size           &   256   &   256  & 256 & 256 & 256 & 256\\
Gradient Clipping    &  0.1    &   0.1   & 0.1 & 0.1 & 0.1 & 0.1               \\
\bottomrule
\end{tabular}              
\end{center}
\end{table}

\end{document}